\documentclass[10pt,twocolumn,letterpaper]{article}
\usepackage[pagenumbers]{cvpr}
\usepackage{stfloats}
\usepackage[utf8]{inputenc}
\usepackage[T1]{fontenc}
\usepackage{algorithm}
\usepackage{algorithmic}
\usepackage{array}
\usepackage{textcomp}
\usepackage{url}
\usepackage{microtype}
\newcommand{\cmark}{\checkmark}
\newcommand{\xmark}{--}

\definecolor{cvprblue}{rgb}{0.21,0.49,0.74}
\usepackage[pagebackref,breaklinks,colorlinks,allcolors=cvprblue]{hyperref}

\def\confName{CVPR}
\def\confYear{2027}
\title{SRP: Anticipating Driving Failures from Structured Policy Representations}
\author{Jiawei Liang \quad Tianrui Lou \quad Le Yang \quad Xianghao Jiao \\
Mingbo Yang \quad Wei Wang \quad Xiaochun Cao \\
Sun Yat-sen University}
\hypersetup{pdftitle={SRP: Anticipating Driving Failures from Structured Policy Representations},pdfauthor={Jiawei Liang, Tianrui Lou, Le Yang, Xianghao Jiao, Mingbo Yang, Wei Wang, Xiaochun Cao},pdfsubject={}}

\begin{document}
\maketitle

\begin{abstract}
End-to-end driving policies form structured internal representations of traffic, ego behavior, and road constraints. These representations offer a monitoring interface for anticipating failures, complementing risk signals from trajectories, controls, and manually designed features. We introduce the Structured Representation Probe (SRP), a lightweight monitor that reads actor, ego, and map representations from a frozen policy. SRP learns a separate linear risk direction for each source and aggregates element scores through cardinality-normalized log-mean-exp pooling, retaining sensitivity to salient evidence while reducing dependence on token count. Event-window supervision links the readout to upcoming hard failures, and auxiliary event type heads jointly train the source directions across failure modes. Online alerting uses a single generic score without modifying the driving policy. Bench2Drive experiments on DriveTransformer, UniAD, and VAD show higher frame-level PR-AUC and ROC-AUC than the evaluated runtime baselines. On DriveTransformer at a horizon of 2 seconds, ROC-AUC reaches 0.833 versus 0.715 for temporal consistency, with 2.67\% readout runtime overhead. The trained probes also retain the ability to distinguish impending failures on new routes.
\end{abstract}

\section{Introduction}

Modern end-to-end autonomous driving models increasingly maintain structured internal representations of surrounding actors, the ego vehicle, and road geometry. These representations support interaction reasoning and planning within the driving policy \cite{jia2025drivetransformer,hu2023planning,jiang2023vad,liao2022maptr,jia2023thinktwice,chen2024vadv2}. A deployed policy also needs timely warnings of impending failures. The internal computation used to select driving actions offers a complementary source of evidence for this task.

\begin{figure}[t]
\centering
\includegraphics[width=\linewidth]{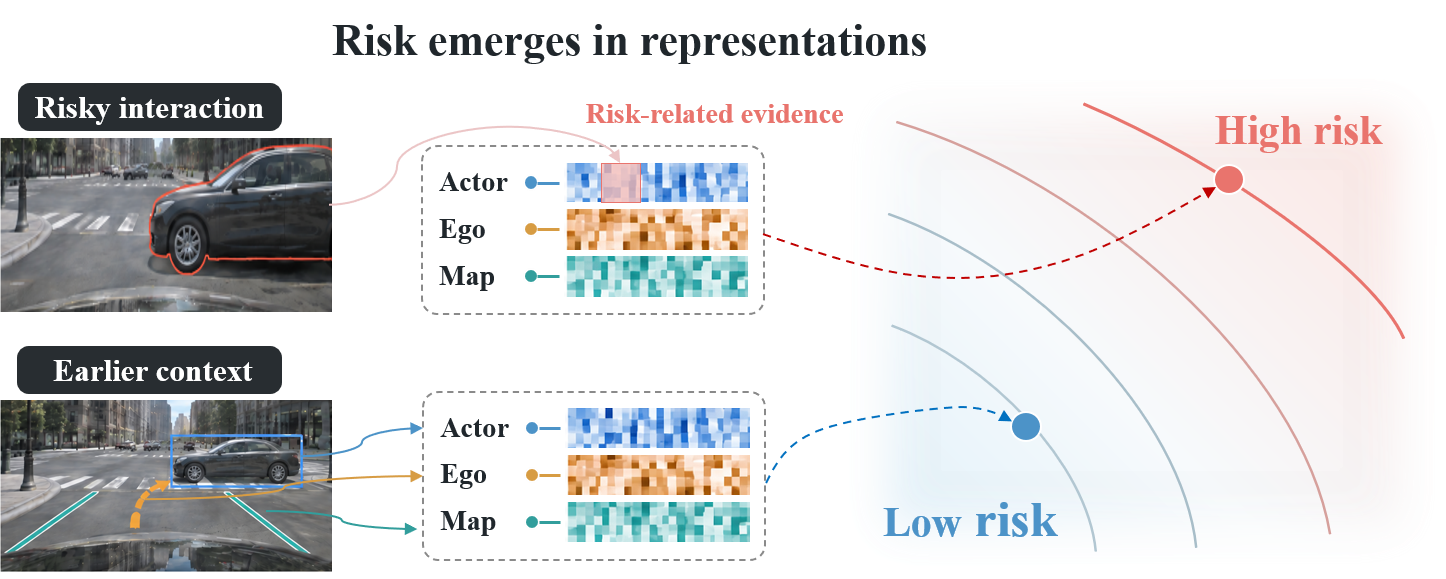}
\caption{Our key insight is that an existing end-to-end driving policy already carries early safety cues in its internal representations, making those representations a natural substrate for risk anticipation.}
\label{fig:main_overview}
\end{figure}

Failure anticipation asks whether information available at the current policy step predicts a hard failure within a short future window, rather than merely identifying an unusual scene. Runtime monitors draw on trajectories, control stability, uncertainty, and rule compliance \cite{hecker2018failure,kuhn2020introspective,antonante2023taskaware,farid2022taskrelevant,stocco2019misbehaviour,mehlhorn2023ruling,lee2020identifying,manas2023legal,michelmore2018evaluating,michelmore2019uncertainty}. We study a complementary interface: the internal computation through which the policy relates scene information to intended behavior (Figure~\ref{fig:main_overview}). Reusing this computation may support early warning without adding another perception or planning stack; it does not presume that output-based signals lack useful failure information.

Two properties of this interface guide the readout. First, actor, ego, and map representations serve different decision-making roles and can occupy different feature spaces \cite{hu2023planning,jiang2023vad,jia2025drivetransformer}, motivating source-specific projections. Second, salient evidence may be concentrated in a few tokens. Averaging can dilute such evidence, while unnormalized aggregation can respond to token count alone. We therefore retain source identity during projection and normalize smooth evidence pooling by set cardinality.

We introduce the Structured Representation Probe (SRP), which reads these interfaces from frozen driving policies \cite{jia2025drivetransformer,hu2023planning,jiang2023vad}. Separate linear risk directions project source tokens into element logits; cardinality-normalized log-mean-exp pooling combines them into a generic warning. Closed-loop event windows provide supervision \cite{chan2016anticipating,jia2024bench2drive}. Auxiliary event-type heads further train the shared directions, but online alerting needs only the generic score. This design couples the policy's semantic organization with a compact readout, keeping warning prediction separate from driving.

Bench2Drive experiments \cite{jia2024bench2drive} cover DriveTransformer, UniAD, and VAD. SRP improves frame-level PR-AUC and ROC-AUC over the evaluated runtime baselines and has the highest pLead at diagnostic thresholds selected under a 5\% FPR constraint on evaluation routes. We separately examine thresholds selected on validation routes and held fixed: event-level advantages then depend on the policy, with temporal consistency outperforming SRP on DriveTransformer event recall and pLead. Logistic-regression baseline variants and source ablations further delimit the comparison. The same probes retain discrimination on new routes without updating the policy.

Our contributions are threefold:
\begin{itemize}
  \item We formulate failure anticipation as lightweight monitoring of a frozen driving policy through its structured actor, ego, and map representation interfaces.
  \item We propose SRP, which combines separate linear risk directions for each source, cardinality-normalized evidence pooling, and auxiliary supervision by event type in a single generic warning monitor.
  \item We evaluate SRP across driving policies and evaluation settings. Results show improved frame discrimination and warning lead over the evaluated runtime monitors.
\end{itemize}

\section{Related Work}

\noindent\textbf{End-to-end driving representations.}
Driving architectures have evolved from navigation-conditioned sensor-to-control mappings \cite{codevilla2018conditional,hawke2019urban} toward structured interfaces for perception, interaction, and planning. UniAD uses a planning-oriented query interface \cite{hu2023planning}; VAD organizes agents and maps as vectorized representations \cite{jiang2023vad}; and DriveTransformer maintains interacting agent, map, and planning queries \cite{jia2025drivetransformer}. MapTR similarly exposes structured road elements \cite{liao2022maptr}. Rather than redesign these policies, SRP attaches a warning readout to their existing actor, ego, and map representations.

\noindent\textbf{Failure anticipation and runtime monitoring.}
Learned introspection predicts failures from visual observations and vehicle state \cite{hecker2018failure,kuhn2020introspective}. Other monitors examine predictive uncertainty \cite{michelmore2018evaluating,michelmore2019uncertainty} or unsupported execution contexts \cite{stocco2019misbehaviour}. Task-aware and task-relevant approaches relate traffic configurations and prediction errors to downstream planning costs \cite{antonante2023taskaware,farid2022taskrelevant,yang2026attribution}. SRP instead learns failure evidence from structured policy representations, retaining a single generic alarm. Our experiments compare adapted runtime monitors and additional logistic-regression variants; they do not establish a universal ranking of internal versus output-based information.

\noindent\textbf{Representation probing.}
Restricted linear probes test whether target information is accessible from intermediate activations \cite{alain2016understanding}. Concept-oriented analyses relate units or directions to visual semantics \cite{bau2017network,kim2018interpretability,liang2026mllm}, while SVCCA and CKA compare representation structure \cite{raghu2017svcca,kornblith2019similarity,jiao2026attribution}. Control tasks caution against attributing a probe's learning ability entirely to its inputs \cite{hewitt2019designing}. We adopt a lightweight readout for a temporal target: proximity to recorded driving failures. Source, supervision, and readout comparisons examine this design without treating diagnostic source scores as causal explanations.

\section{Method}

SRP attaches a lightweight warning readout to the structured representation interface of a frozen driving policy. Its design combines separate risk directions for each source, cardinality-normalized aggregation, and auxiliary supervision by event type. The source directions retain the actor, ego, and map organization, while the generic aggregation produces the score used for online alerting.

\subsection{Monitoring Formulation}

Let $f_\theta$ be an end-to-end driving policy. At time $t$, the policy maps the observation history $o_{\le t}$ to a driving plan or control $u_t$. A fixed exporter $\Phi_\theta$ reads the structured internal representations exposed by the policy:
\begin{equation}
u_t = f_\theta(o_{\le t}), \qquad z_t = \Phi_\theta(o_{\le t}).
\end{equation}
SRP reads from this interface without modifying $f_\theta$ and maps the representations to a scalar generic warning score,
\begin{equation}
m_\phi(z_t) = r_t,
\end{equation}
where $r_t\in[0,1]$ is the score used for online alerting. Auxiliary event type heads produce scores $r_{t,c}\in[0,1]$ for the corresponding event-window labels. They jointly supervise the representation readout during training and provide diagnostic scores for each event type.

\begin{figure*}[t]
\centering
\includegraphics[width=\textwidth]{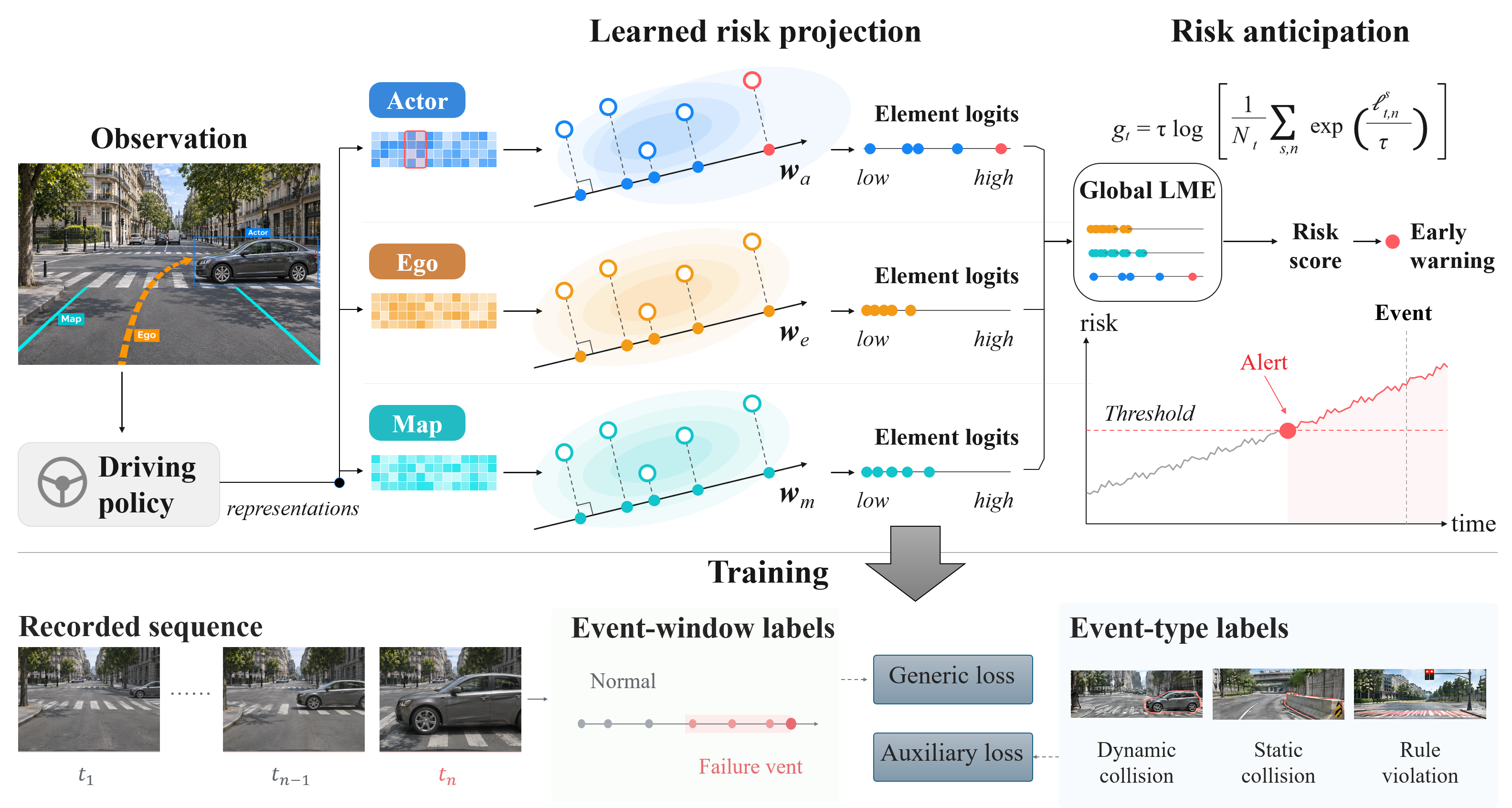}
\caption{SRP framework. Frozen-policy actor, ego, and map tokens yield element logits. Both branches use cardinality-normalized LME. Event-window labels supervise the generic readout and three event heads; $\mathcal{L}_{\mathrm{type}}$ is their mean weighted BCE. Flames mark trainable risk directions; online alerting uses only $r$ with a validation-selected threshold.}
\label{fig:framework}
\end{figure*}

The monitor reuses the policy's internal computation during closed-loop execution. Figure~\ref{fig:framework} shows the generic alerting path and its training with event-window labels and auxiliary heads.

\subsection{Structured Driving Evidence}

Driving decisions must account for surrounding traffic, the ego vehicle's intended maneuver, and road constraints. These considerations motivate three complementary sources of evidence for failure anticipation: actor, ego, and map representations. We organize the exported representations by source as
\begin{equation}
z_t \equiv \{A_t, E_t, M_t\},
\end{equation}
where $A_t$, $E_t$, and $M_t$ denote evidence organized around actors, ego behavior, and maps, respectively. This notation describes source roles rather than a particular tensor name, so it can be instantiated across driving models.

\begin{equation}
A_t = \{a_{t,i}\}_{i=1}^{N_t^a},\quad
E_t = \{e_{t,k}\}_{k=1}^{N_t^e},\quad
M_t = \{m_{t,j}\}_{j=1}^{N_t^m},
\end{equation}
where $a_{t,i}\in\mathbb{R}^{d_a}$, $e_{t,k}\in\mathbb{R}^{d_e}$, and $m_{t,j}\in\mathbb{R}^{d_m}$.

Actor tokens describe surrounding participants and their interactions with the ego vehicle \cite{hu2023planning,jiang2023vad,jia2025drivetransformer}. Ego tokens represent the policy's motion or planning state \cite{hu2023planning,jia2025drivetransformer}, while map tokens encode road geometry and semantic constraints \cite{liao2022maptr,jiang2023vad}. These source roles are shared across policies, but their tensor dimensions and organization need not be identical. SRP uses a separate projection for each source rather than requiring a common embedding dimension.

Each evidence source is assigned its own risk direction. Let $\mathcal{S}=\{a,e,m\}$ and let $X_t^s=\{x_{t,n}^s\}_{n=1}^{N_t^s}$ denote one of the three evidence sets. After normalization within each source, an instance contributes risk evidence through
\begin{equation}
\label{eq:srp_element_logits}
\ell_{t,n}^{s}=\langle w_s,\tilde{x}_{t,n}^{s}\rangle+b_s,
\qquad s\in\mathcal{S}.
\end{equation}
Each $w_s$ operates in its source's representation space, preserving the distinction between actor interactions, ego behavior, and road constraints.

\subsection{Risk Scoring by Source}

A scene may contain many representation elements but only a few strong risk signals, such as a conflicting actor or a critical road constraint. Average pooling can dilute these signals \cite{zaheer2017deep,qi2017pointnet,ilse2018attention}, while unnormalized log-sum-exp increases when the same collection of logits is repeated. Drawing on smooth pooling operators used in multiple-instance learning \cite{cohen2016deep,carbonneau2018multiple,fatima2023comprehensive}, we use cardinality-normalized log-mean-exp pooling to summarize each source:
\begin{equation}
\label{eq:srp_source_pooling}
\rho_t^s =
\tau \log\left(
\frac{1}{N_t^s}
\sum_{n=1}^{N_t^s}
\exp(\ell_{t,n}^{s}/\tau)
\right),
\qquad s\in\{a,e,m\}.
\end{equation}
The temperature $\tau$ controls the sharpness of the evidence readout. This aggregation remains sensitive to instances indicating high risk while reducing dependence on the number of actor, ego, or map elements available in a frame.

\noindent\textbf{Set function property.}
Because the pooled score acts on an unordered set of representation vectors, it is permutation invariant in element order and behaves like a smooth max operator \cite{zaheer2017deep,kimura2024permutation,tang2021sensory}. For any nonempty source bag and $\tau>0$, it satisfies
\begin{equation}
\bar{\ell}_t^s \le \rho_t^s \le \max_{n}\ell_{t,n}^s,
\qquad
\lim_{\tau\rightarrow 0^+}\rho_t^s = \max_{n}\ell_{t,n}^s,
\end{equation}
where $\bar{\ell}_t^s = \frac{1}{N_t^s}\sum_{n=1}^{N_t^s}\ell_{t,n}^s$. Repeating the entire collection of logits leaves this normalized score unchanged. Adding or removing particular elements changes the exponential mean, so the readout remains sensitive to set composition as well as salient high values.

The generic warning used for inference aggregates the instance evidence across all sources:
\begin{equation}
\label{eq:srp_generic_pooling}
g_t =
\tau \log\left(
\frac{1}{N_t}
\sum_{s\in\{a,e,m\}}\sum_{n=1}^{N_t^s}
\exp(\ell_{t,n}^{s}/\tau)
\right),\;
r_t=\sigma(g_t).
\end{equation}
Here $N_t=\sum_{s\in\mathcal{S}}N_t^s$. Generic pooling weights each source by its token proportion; sufficiently large element logits can dominate the exponential sum. Replicating all source bags by the same factor leaves $g_t$ unchanged, but changing one source's cardinality or token composition need not. Source-normalized pooling therefore does not imply source-balanced generic aggregation. During training, each auxiliary event-type head reads the vector of pooled source scores:
\begin{equation}
\label{eq:srp_typed_logits}
h_{t,c}=\alpha_c^\top[\rho_t^a,\rho_t^e,\rho_t^m]+\beta_c,
\qquad
r_{t,c}=\sigma(h_{t,c}).
\end{equation}
Each auxiliary head is a linear readout of the three pooled source scores. Its loss trains the shared source directions with supervision by event type, and its output provides a diagnostic score for the corresponding event label.

To train the generic and event-type readouts, we construct temporal labels from recorded hard failures \cite{chan2016anticipating,jia2024bench2drive}. For each route $r$, let
\begin{equation}
\mathcal{T}_r = \{t_r^{(1)}, \dots, t_r^{(K_r)}\}
\end{equation}
be the set of frames containing hard failures, matched to policy steps. For auxiliary supervision, the covered event types are grouped into
\begin{equation}
\mathcal{C}=\{\mathrm{\mathbf{D}ynamic},\mathrm{\mathbf{S}tatic},\mathrm{\mathbf{R}ule}\},
\end{equation}
where dynamic collision includes vehicle and pedestrian collisions, static collision includes collisions with static objects, and rule violation includes red light and stop sign infractions and route deviation. Let $\mathcal{T}_{r,c}$ denote the event frames of type $c$ on route $r$. Given an anticipation window $W$, the generic and event type labels use the same window rule:
\begin{equation}
\begin{aligned}
y_t^{(W)}
&=
\mathbf{1}\left[\exists t^\star \in \mathcal{T}_r:
0 \le t^\star - t \le W\right],\\
y_{t,c}^{(W)}
&=
\mathbf{1}\left[\exists t^\star \in \mathcal{T}_{r,c}:
0 \le t^\star - t \le W\right],
\end{aligned}
\end{equation}
Positive labels are constructed from the union of the windows for target events, each extending backward from and including its event frame. A frame following one event can therefore remain positive if it falls within the window of a later target event.

SRP is trained with weighted binary cross-entropy:
\begin{equation}
\label{eq:srp_loss}
\mathcal{L}
=
\mathrm{BCE}(g_t,y_t^{(W)})
+
\lambda
\frac{1}{|\mathcal{C}|}
\sum_{c\in\mathcal{C}}
\mathrm{BCE}(h_{t,c},y_{t,c}^{(W)}).
\end{equation}
Class weights compensate for the rarity of positive event-window frames, and $\lambda$ controls the strength of the mean event type supervision term.

Online inference reuses the saved feature statistics and probe parameters and compares $r_t$ with a generic threshold selected on the validation split. The source scores $(\rho_t^a,\rho_t^e,\rho_t^m)$ and auxiliary scores $\{r_{t,c}\}_{c\in\mathcal{C}}$ are optional diagnostic outputs; alerting uses only $r_t$.

Training and online inference are summarized in Supplementary Sec.~C using the same notation.

\begin{table*}[!t]
\caption{Runtime-monitor comparison on Bench2Drive test routes (2-s windows). Diagnostic thresholds are selected on the test split under empirical frame-level FPR$\le$5\%.}
\label{tab:baseline_comparison}
\centering
\small
\setlength{\tabcolsep}{2.5pt}
\begin{tabular*}{\textwidth}{@{\extracolsep{\fill}}llccccc@{}}
\toprule
Model & Method & PR-AUC & ROC-AUC & $R_F$@5\%FPR & $R_E$@5\%FPR & pLead@5\%FPR (s) \\
\midrule
DriveTransformer & Temporal Consistency & 0.127 & 0.715 & 0.145 & \textbf{0.505} & 0.633 \\
DriveTransformer & Task-Aware Risk & 0.088 & 0.570 & 0.060 & 0.431 & 0.532 \\
DriveTransformer & TRFD & 0.080 & 0.496 & 0.046 & 0.320 & 0.337 \\
DriveTransformer & Ours & \textbf{0.261} & \textbf{0.833} & \textbf{0.300} & 0.497 & \textbf{0.701} \\
\midrule
UniAD & Temporal Consistency & 0.167 & 0.751 & 0.175 & 0.573 & 0.768 \\
UniAD & Task-Aware Risk & 0.070 & 0.506 & 0.042 & 0.339 & 0.455 \\
UniAD & TRFD & 0.090 & 0.520 & 0.094 & 0.489 & 0.653 \\
UniAD & Ours & \textbf{0.339} & \textbf{0.868} & \textbf{0.439} & \textbf{0.725} & \textbf{1.124} \\
\midrule
VAD & Temporal Consistency & 0.131 & 0.810 & 0.280 & 0.723 & 1.159 \\
VAD & Task-Aware Risk & 0.055 & 0.677 & 0.109 & \textbf{0.739} & 0.870 \\
VAD & TRFD & 0.060 & 0.639 & 0.145 & 0.667 & 0.847 \\
VAD & Ours & \textbf{0.320} & \textbf{0.853} & \textbf{0.559} & 0.713 & \textbf{1.229} \\
\bottomrule
\end{tabular*}
\end{table*}

\section{Experiments}

We ask how well structured representations support warning, which sources and supervision contribute, and how performance transfers to new routes. Main results use evaluation-set-selected diagnostic thresholds; we separately report validation-fixed thresholds and logistic-regression baseline variants in Supplementary Secs.~A and B. Implementation, further quantitative diagnostics, and additional qualitative analyses appear in Supplementary Secs.~C--E.

\subsection{Experimental Setup}

\noindent\textbf{Dataset.}
We use Bench2Drive \cite{jia2024bench2drive}, a closed-loop benchmark in CARLA \cite{dosovitskiy2017carla} with 220 routes covering multiple towns and weather conditions. Each driving policy runs on the same benchmark routes. We collect its internal representations online and pair them with hard failures recorded during its own executions.

The target events are collisions, red light infractions, stop sign infractions, and route deviations. Auxiliary supervision groups them into dynamic collision, static collision, and rule violation, with the latter including red light, stop sign, and route deviation. Generic positive labels are the union of all windows for target events, including the event frames. Unless otherwise stated, training and evaluation use 40 simulator steps (2 seconds). Rollouts may terminate on collision, route deviation, or blocked criteria; red light and stop sign violations are recorded but do not themselves terminate a rollout.

\noindent\textbf{Models and Baselines.}
Our driving models are DriveTransformer \cite{jia2025drivetransformer}, UniAD \cite{hu2023planning}, and VAD \cite{jiang2023vad}. Exporters for each model map its internal representations to the common actor, ego, and map interface; the driving policies remain frozen.

We adapt three monitoring approaches to the available policy outputs and the common evaluation protocol applied at each frame. Temporal Consistency (TC), inspired by MomAD \cite{song2025momad}, summarizes changes in recent ego plans and controls. Task-Aware Risk (TAR) \cite{antonante2023taskaware} uses nearby actors' states and trajectories to sample alternative traffic configurations and assess their effect on the current ego plan. Task-Relevant Failure Detection (TRFD) \cite{farid2022taskrelevant} takes the ego plan and predicted actor modes, forms nominal and alternative scenes, and compares the associated planning costs. Each implementation emits one risk score per frame from the corresponding policy under the same closed-loop protocol.

\noindent\textbf{Training Details.}
All policies use disjoint 132/44/44 train/validation/test route splits. Probes are trained for 50 epochs with AdamW \cite{loshchilov2019decoupled}, learning rate $10^{-3}$, weight decay $10^{-4}$, batch size 128, and weighted binary cross-entropy. Source feature statistics are estimated on training routes and reused for validation, testing, and transfer. The default pooling temperature is $\tau=1$, and full SRP uses auxiliary weight $\lambda=0.5$. Generic validation PR-AUC selects the checkpoint. We save the probe, training statistics, and source-validation threshold for online use; the full procedure is in Supplementary Sec.~C.

\noindent\textbf{Evaluation Metrics.}
We report frame-level PR-AUC \cite{saito2015precision}, ROC-AUC \cite{fawcett2006introduction}, frame recall ($R_F$), event recall ($R_E$), and penalized lead time (pLead, seconds). $R_F$ covers positive frames. $R_E$ is the fraction of selected events with an alert strictly before the event and within the evaluation window; each positive route contributes its first event of the evaluated label. pLead averages the earliest qualifying alert's lead over these events, assigning zero to misses. Thus it reflects both coverage and timing. Formal definitions and threshold protocols are in Supplementary Sec.~A.

\begin{table*}[!t]
\caption{Retrained DriveTransformer probe ablations (2-s windows, test-selected diagnostic thresholds). Checks mark enabled sources and auxiliary supervision.}
\label{tab:ablation_plan}
\centering
\small
\setlength{\tabcolsep}{2pt}
\begin{tabular*}{\textwidth}{@{\extracolsep{\fill}}lccccccccc@{}}
\toprule
Variant & Actor & Ego & Map & Aux. Sup. & PR-AUC & ROC-AUC & $R_F$@5\%FPR & $R_E$@5\%FPR & pLead@5\%FPR (s) \\
\midrule
No actor & \xmark & \cmark & \cmark & \cmark & 0.250 & 0.813 & 0.291 & 0.506 & 0.712 \\
No ego & \cmark & \xmark & \cmark & \cmark & 0.229 & 0.784 & 0.261 & 0.310 & 0.504 \\
No map & \cmark & \cmark & \xmark & \cmark & 0.251 & 0.821 & 0.289 & 0.485 & 0.680 \\
\midrule
Actor only & \cmark & \xmark & \xmark & \cmark & 0.239 & 0.812 & 0.286 & 0.394 & 0.561 \\
Ego only & \xmark & \cmark & \xmark & \cmark & 0.245 & 0.795 & 0.285 & 0.495 & 0.704 \\
Map only & \xmark & \xmark & \cmark & \cmark & 0.196 & 0.784 & 0.204 & 0.372 & 0.516 \\
\midrule
Generic only & \cmark & \cmark & \cmark & \xmark & 0.224 & 0.793 & 0.248 & 0.424 & 0.561 \\
\midrule
Full & \cmark & \cmark & \cmark & \cmark & 0.261 & 0.833 & 0.300 & 0.497 & 0.701 \\
\bottomrule
\end{tabular*}
\end{table*}

\begin{figure}[!t]
\centering
\includegraphics[width=\columnwidth]{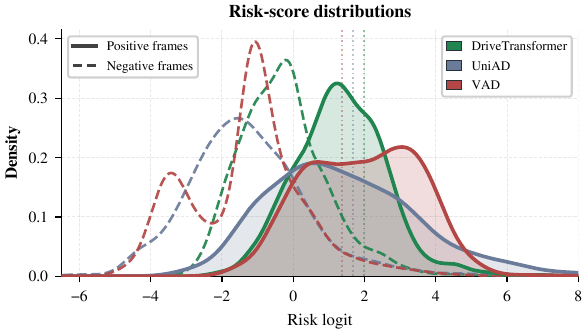}
\caption{Risk logits on Bench2Drive test routes. Solid/dashed curves: positive/negative frames. Dotted lines: mean policy-specific test-selected thresholds (frame FPR$\le$5\%).}
\label{fig:risk_logit_dist}
\end{figure}

\begin{figure}[!t]
\centering
\includegraphics[width=\columnwidth]{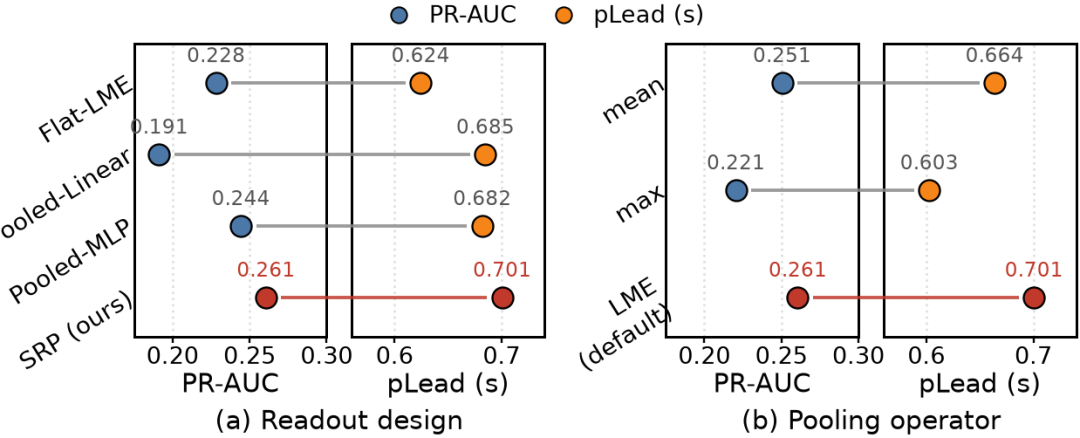}
\caption{Readout and pooling comparisons on DriveTransformer. Paired markers show mean PR-AUC and pLead; connecting lines identify the same setting. Red markers highlight SRP in (a) and its default LME pooling in (b).}
\label{fig:readout_pooling}
\end{figure}

\begin{figure*}[!t]
\centering
\includegraphics[width=\textwidth]{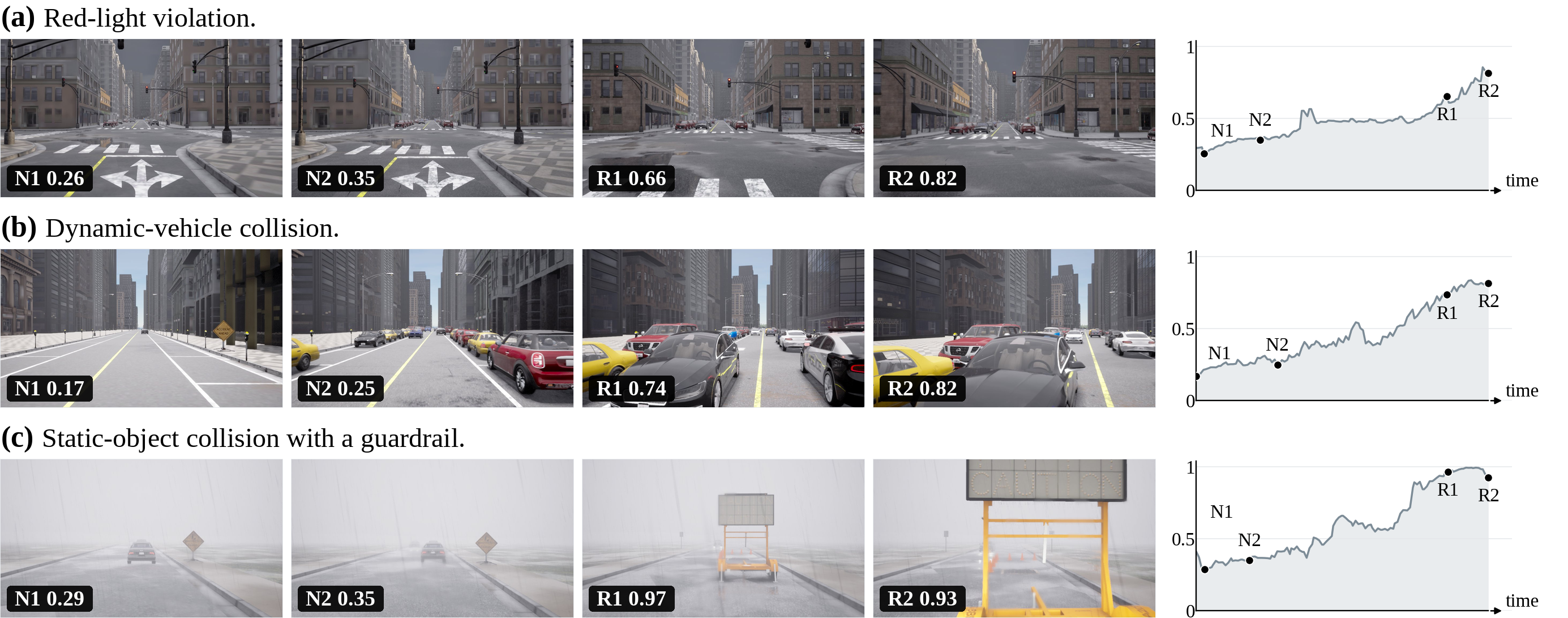}
\caption{Representative qualitative warning cases. SRP raises its generic risk score before three distinct hard failures: a red light violation, a collision with another vehicle, and a collision with a roadside guardrail.}
\label{fig:representative_cases}
\end{figure*}

\subsection{Main Benchmark Results}

Table~\ref{tab:baseline_comparison} compares SRP with runtime monitors \cite{song2025momad,antonante2023taskaware,farid2022taskrelevant} on Bench2Drive \cite{jia2024bench2drive}, using test-selected diagnostic thresholds under empirical FPR$\le$5\%. Validation-selected thresholds applied unchanged to test are reported separately in Supplementary Sec.~A.

At the diagnostic operating points, SRP leads PR-AUC, ROC-AUC, and $R_F$ across all three policies. On DriveTransformer \cite{jia2025drivetransformer}, ROC-AUC rises from 0.715 for temporal consistency to 0.833, while pLead increases from 0.633 to 0.701 seconds with nearly unchanged event recall (0.505 versus 0.497). On UniAD \cite{hu2023planning}, SRP improves both coverage and penalized lead, reaching $R_E=0.725$ and pLead of 1.124 seconds. On VAD \cite{jiang2023vad}, SRP has the highest pLead at 1.229 seconds, whereas Task-Aware Risk has the highest event recall (0.739 versus 0.713). Ranking quality, event coverage, and warning lead therefore provide complementary evidence rather than interchangeable measures.

\noindent\textbf{Risk-score separation.}
Figure~\ref{fig:risk_logit_dist} shows positive event-window frames shifted toward higher logits for all three policies, with substantial overlap remaining. The dotted thresholds locate the high-risk tail used for diagnostic alerting, visually complementing the ranking results in Table~\ref{tab:baseline_comparison}. Since each probe is fitted separately, absolute logit magnitudes are not calibrated across policies; the evidence is within-policy separation, not a universal risk scale.

\noindent\textbf{Validation-fixed benchmark.}
With source validation thresholds fixed on test (Supplementary Sec.~A), SRP leads event recall and pLead on UniAD and VAD. On DriveTransformer, TC instead leads $R_E$ (0.601 versus 0.437) and pLead (0.705 versus 0.619 seconds), at observed FPRs of 4.30\% and 5.20\%. Fixed-threshold results are not equal-test-FPR comparisons and need not retain the diagnostic ordering.

\subsection{Source, Supervision, and Readout Analysis}

Table~\ref{tab:ablation_plan} separates source selection from auxiliary supervision. Source variants are retrained with different representation subsets; the supervision variant retains all sources but removes the event type loss. All variants use DriveTransformer \cite{jia2025drivetransformer} with the same windows of 2 seconds, route split, labels, and metric protocol as the main comparison.

\noindent\textbf{Source selection.}
All three sources together achieve the highest PR-AUC and ROC-AUC among the retrained variants. The ego source carries substantial warning evidence: pLead of the probe using only ego representations is close to that of the full probe (0.704 versus 0.701 seconds), and ego removal produces the largest decline among the variants using two sources, reducing $R_E$ from 0.497 to 0.310. Removing actor slightly increases event recall and pLead at the selected operating point. Thus the sources contribute differently to ranking and thresholded warning, with their combination giving the strongest overall frame discrimination.

\noindent\textbf{Auxiliary supervision.}
Auxiliary event supervision improves all five reported metrics for the generic warning with the source set held fixed. Adding the event-type loss raises PR-AUC from 0.224 to 0.261 and pLead from 0.561 to 0.701 seconds. Since inference uses only the generic score, this comparison measures a training contribution to the shared source directions rather than an ensemble of event-head alarms.

We next compare readout structure and aggregation on the same DriveTransformer representations and 2-second windows. Additional source-deletion and sensitivity results are in Supplementary Sec.~D.

\noindent\textbf{Readout design.}
Figure~\ref{fig:readout_pooling}(a) compares four ways to read the same policy representations, following restricted representation probing \cite{alain2016understanding}. Flat-LME shares a linear direction across sources. Pooled-Linear classifies concatenated source-wise max-pooled features, whereas Pooled-MLP uses mean-pooled features and a hidden layer of width 16. SRP has the highest mean PR-AUC and pLead (0.261 and 0.701 seconds). Because the alternatives change both feature summarization and classifier structure, this is a readout-design comparison, not an isolated test of model capacity. It supports the tested source-specific readout without implying that a larger classifier could never improve these representations.

\noindent\textbf{Pooling.}
We retrain SRP with mean, max, or LME aggregation in both the source-wise and generic pooling branches. LME achieves the highest mean PR-AUC and pLead among the three choices (Figure~\ref{fig:readout_pooling}(b)), increasing them from 0.251 to 0.261 and from 0.664 to 0.701 seconds relative to mean pooling.

\noindent\textbf{Logistic-regression baseline variants.}
Supplementary Sec.~B uses the same test-selected diagnostic protocol. TC-LogReg improves PR-AUC over original TC across all three policies. SRP full retains higher PR-AUC and ROC-AUC, but its pLead margins on DriveTransformer and VAD are small (0.014 and 0.026 seconds), and DriveTransformer event recall is lower (0.497 versus 0.528). These differences are descriptive, not significance claims.

DriveTransformer generic-only SRP has PR-AUC of 0.224 and pLead of 0.561 seconds, versus 0.231 and 0.687 seconds for TC-LogReg. The controlled generic-only/full ablation supports auxiliary supervision, but the cross-method comparison also changes inputs and readouts. It does not isolate representation information or a fully supervision-matched advantage.

\begin{figure}[!t]
\centering
\includegraphics[width=\linewidth]{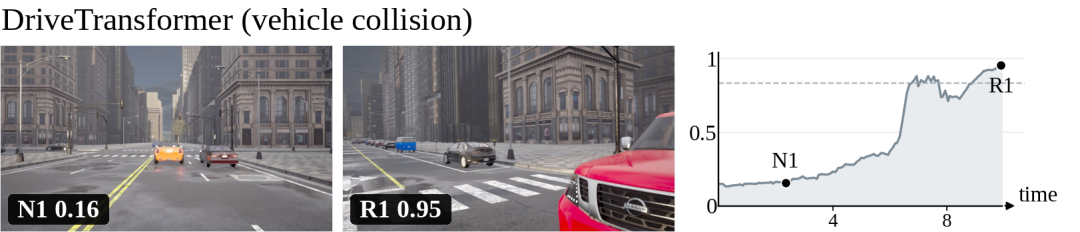}\par\vspace{3pt}
\includegraphics[width=\linewidth]{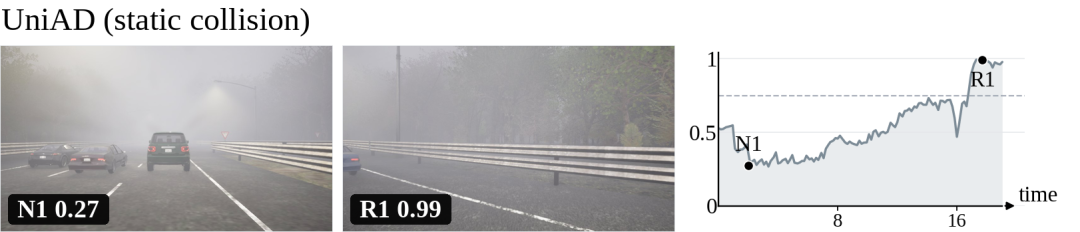}\par\vspace{3pt}
\includegraphics[width=\linewidth]{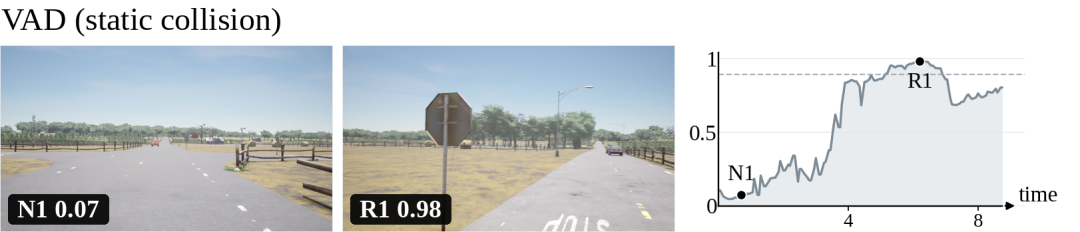}
\caption{Qualitative warning cases across policies: a vehicle collision for DriveTransformer and static collisions for UniAD and VAD. N1 and R1 link snapshots to generic score traces.}
\label{fig:policy_warning_cases}
\end{figure}

\subsection{Transfer and Source Reliance}

\noindent\textbf{Transfer to new routes.}
We evaluate probes trained on the original Bench2Drive \cite{jia2024bench2drive} routes on two targets: the new Bench2Drive evaluation set with 220 routes and the 20 validation routes of CARLA Leaderboard 2.0 (LB2) \cite{carlaLeaderboard20}. Policy and probe parameters, together with the training feature statistics, remain fixed. Both targets use generic labels and windows of 2 seconds. For diagnostic comparison under a common FPR constraint, each target uses a separately selected threshold under the same 5\% frame-level FPR rule, distinct from the source validation threshold saved for deployment.

The probes trained on the original routes retain frame-level discrimination on both targets (Table~\ref{tab:data_transfer}), with ROC-AUC ranging from 0.807 to 0.923 on the new Bench2Drive routes and from 0.769 to 0.878 on LB2. Event coverage varies more across policies on LB2: $R_E$ ranges from 0.444 to 0.765, compared with 0.681 to 0.793 on the new Bench2Drive set. For UniAD \cite{hu2023planning}, PR-AUC is higher on LB2 (0.380 versus 0.324), while event recall is lower (0.444 versus 0.705). On LB2, UniAD has the highest PR-AUC but the lowest event recall, while VAD \cite{jiang2023vad} leads ROC-AUC, event recall, and pLead. This contrast shows why frame ranking, event coverage, and warning lead should be considered together when assessing generalization.

\begin{table*}[!t]
\centering
\caption{Transfer to new Bench2Drive routes (220) and LB2 validation routes (20). Policy and probe are fixed; diagnostic thresholds are reselected on each target under empirical frame-level FPR$\le$5\%.}
\label{tab:data_transfer}
\small
\setlength{\tabcolsep}{3pt}
\begin{tabular*}{\textwidth}{@{\extracolsep{\fill}}llccccc@{}}
\toprule
Target data & Model & PR-AUC & ROC-AUC & $R_F$@5\%FPR & $R_E$@5\%FPR & pLead@5\%FPR (s) \\
\midrule
Bench2Drive (new) & DriveTransformer & 0.363 & 0.807 & 0.360 & 0.681 & 1.029 \\
Bench2Drive (new) & UniAD & 0.324 & 0.864 & 0.526 & 0.705 & 1.138 \\
Bench2Drive (new) & VAD & 0.314 & 0.923 & 0.604 & 0.793 & 1.361 \\
\midrule
LB2 & DriveTransformer & 0.312 & 0.769 & 0.344 & 0.737 & 1.142 \\
LB2 & UniAD & 0.380 & 0.829 & 0.456 & 0.444 & 0.683 \\
LB2 & VAD & 0.371 & 0.878 & 0.513 & 0.765 & 1.406 \\
\bottomrule
\end{tabular*}
\end{table*}

\begin{table}[t]
\caption{Fixed source-validation thresholds on new Bench2Drive and LB2 routes. FPR is the observed false-positive rate.}
\label{tab:fixed_threshold_transfer}
\centering
\small
\setlength{\tabcolsep}{3pt}
\begin{tabular*}{\columnwidth}{@{\extracolsep{\fill}}llccc@{}}
\toprule
Evaluation set & Policy & FPR (\%) & $R_E$ & pLead (s) \\
\midrule
B2D new & DT & 6.36 & 0.723 & 1.157 \\
B2D new & UniAD & 5.53 & 0.719 & 1.181 \\
B2D new & VAD & 2.66 & 0.690 & 1.073 \\
\midrule
LB2 & DT & 2.88 & 0.684 & 1.079 \\
LB2 & UniAD & 5.56 & 0.500 & 0.728 \\
LB2 & VAD & 2.78 & 0.647 & 1.159 \\
\bottomrule
\end{tabular*}
\end{table}

\noindent\textbf{Fixed-threshold transfer.}
We keep the thresholds selected on the source validation routes fixed on both targets (Table~\ref{tab:fixed_threshold_transfer}). Observed FPR ranges from 2.66\% to 6.36\% on the new Bench2Drive routes and from 2.78\% to 5.56\% on LB2, with event recall of 0.690--0.723 and 0.500--0.684, respectively. Mean pLead exceeds one second in all settings except UniAD on LB2 (0.728 seconds).

\noindent\textbf{Fixed-probe source deletion.}
Supplementary Sec.~D deletes actor, ego, or map tokens from the fitted DriveTransformer probe on new Bench2Drive routes. Parameters remain fixed, and generic pooling is renormalized over the remaining tokens. Ego deletion reduces PR-AUC from 0.363 to 0.217 and $R_E$ from 0.681 to 0.411; actor deletion has a smaller effect, while map deletion leaves ranking nearly unchanged. Unlike the retraining in Table~\ref{tab:ablation_plan}, this evaluates fitted-source reliance. The datasets differ, and each deletion setting reselects a diagnostic threshold, so the results are neither causal attribution nor fixed-threshold robustness.

\subsection{Qualitative Warning Analysis}

Figure~\ref{fig:representative_cases} links snapshots to generic warning traces for red light, vehicle, and guardrail failures. Figure~\ref{fig:policy_warning_cases} adds cases from DriveTransformer, UniAD, and VAD, with N1 and R1 linking scene snapshots to the traces. These selected cases illustrate warning evolution, not population-level reliability. Event-group lead times and temporal profiles remain in Supplementary Sec.~E; they are distinct from auxiliary-head label evaluation and are not causal explanations.

\subsection{Readout Cost}

\begin{table}[!b]
\centering
\caption{Probe parameters, generic-score FLOPs, and measured readout latency. Percentages are relative to policy runtime.}
\label{tab:overhead}
\small
\begin{tabular*}{\columnwidth}{@{\extracolsep{\fill}}lccc@{}}
\toprule
Model & Params & FLOPs & Overhead \\
\midrule
DriveTransformer & 2.3K & 0.220M & 6.47 ms (2.67\%) \\
UniAD & 0.8K & 0.328M & 1.89 ms (0.20\%) \\
VAD & 0.8K & 0.412M & 3.37 ms (1.58\%) \\
\bottomrule
\end{tabular*}
\end{table}

Table~\ref{tab:overhead} reports parameter counts for the probe alone and the overhead of the generic warning readout. FLOPs exclude the driving policy and auxiliary heads, averaging normalization, token projection, log-mean-exp pooling, and a sigmoid over the recorded frames; one multiply-add counts as two FLOPs. Latency is measured on an NVIDIA A100 GPU for risk score computation, with percentages relative to the policy runtime.

SRP uses 0.8K--2.3K probe parameters, and computation of the generic score takes 0.20\%--2.67\% of policy runtime. DriveTransformer \cite{jia2025drivetransformer} has the lowest FLOP count and the longest measured readout time (6.47 ms), illustrating that arithmetic cost alone does not determine latency. The measured timings establish a modest readout overhead across all three policies.

\section{Discussion}

Structured policy representations provide a useful monitoring interface, but the evidence has distinct scopes. SRP reuses policy computation without changing driving outputs or adding a separate perception stack. Frame discrimination improves across the evaluated policies. Event-level gains depend on both the threshold protocol and the policy: diagnostic lead-time advantages do not uniformly persist with validation-fixed thresholds. Logistic-regression variants strengthen the comparison while showing that auxiliary supervision and readout choices matter; these results do not isolate representation information from supervision or model capacity. Source scores and auxiliary heads are diagnostic associations, not causal explanations or additive components of the alarm. Retraining and source deletion respectively measure attainable performance and fitted reliance (Supplementary Sec.~D). Frame AUC includes event frames, while event recall and pLead require earlier alerts \cite{chan2016anticipating,jia2024bench2drive}. Real-world deployment and alert responses require operational-design-domain validation \cite{mehlhorn2023ruling,lee2020identifying}. Readout timings exclude full integration and do not establish safety certification.

\section{Conclusion}

SRP reads structured actor, ego, and map representations from a frozen driving policy through source-specific linear projections, cardinality-normalized pooling, and auxiliary event supervision. Across the evaluated policies, it improves frame discrimination and warning lead at diagnostic operating points, retains discrimination on new routes, and incurs modest readout cost. Validation-fixed results reveal policy-dependent coverage and lead-time trade-offs. This supports lightweight monitoring through reused policy computation, without conflating diagnostic gains with deployment guarantees.

\clearpage
\section*{LLM Usage Statement}
OpenAI Codex supported prose drafting and revision for the abstract, introduction, related work, experiment descriptions, conclusion, and captions. It also assisted with literature checks, LaTeX typesetting, table layout, and placement of author-provided figures. Experimental measurements and visual examples were drawn from the existing research materials.

{
\small
\bibliographystyle{ieeenat_fullname}
\bibliography{references}
}

\clearpage
\setlength{\textfloatsep}{12pt plus 2pt minus 2pt}
\setlength{\dbltextfloatsep}{12pt plus 2pt minus 2pt}
\setlength{\floatsep}{10pt plus 2pt minus 2pt}
\setlength{\dblfloatsep}{10pt plus 2pt minus 2pt}
\renewcommand{\dblfloatpagefraction}{0.75}
\makeatletter
\newcommand{\suppcontentsentry}[2]{\@dottedtocline{1}{0em}{1.6em}{\hyperref[#1]{#2}}{\pageref{#1}}}
\setlength{\@dblfptop}{0pt}
\setlength{\@dblfpsep}{14pt}
\setlength{\@dblfpbot}{0pt plus 1fil}
\makeatother
\twocolumn[{
\begin{minipage}{\textwidth}
{\centering
\Large
\textbf{\thetitle}\par
\vspace{0.5em}
Supplementary Material\par
\vspace{1.0em}
}
\section*{Contents}
\begingroup
\small
\suppcontentsentry{sec:supp_fixed}{\ref*{sec:supp_fixed}. Evaluation Protocol and Fixed-threshold Results}
\suppcontentsentry{sec:supp_logreg}{\ref*{sec:supp_logreg}. Logistic-regression Baseline Variants}
\suppcontentsentry{sec:supp_implementation}{\ref*{sec:supp_implementation}. Implementation and Training Procedure}
\suppcontentsentry{sec:supp_quantitative}{\ref*{sec:supp_quantitative}. Additional Quantitative Analysis}
\suppcontentsentry{sec:supp_qualitative}{\ref*{sec:supp_qualitative}. Additional Qualitative and Temporal Analysis}
\suppcontentsentry{sec:supp_properties}{\ref*{sec:supp_properties}. Readout Properties and Computational Complexity}
\endgroup
\vspace{10pt}
\end{minipage}
}]
\appendix
\setcounter{table}{0}
\setcounter{figure}{0}
\setcounter{algorithm}{0}
\setcounter{equation}{0}
\renewcommand{\thetable}{S\arabic{table}}
\renewcommand{\thefigure}{S\arabic{figure}}
\renewcommand{\thealgorithm}{S\arabic{algorithm}}
\renewcommand{\theequation}{S\arabic{equation}}

\section{Evaluation Protocol and Fixed-threshold Results}
\label{sec:supp_fixed}

\noindent\textbf{Diagnostic evaluation.}
The main benchmark, retrained ablations, and logistic-regression variants select thresholds on the evaluated test split to maximize frame recall under empirical frame-level FPR$\le$5\%. Each threshold is used for frame recall, event recall, and pLead in that evaluation. Transfer diagnostics select a threshold separately on each target. The notation @5\%FPR identifies this evaluation-set-selected operating point.

\noindent\textbf{Validation-fixed evaluation.}
Table~\ref{tab:validation_fixed_benchmark} applies source-validation thresholds unchanged to the Bench2Drive test routes, with 2-second training and evaluation windows. Its FPR column reports the observed test false-positive rate. PR-AUC and ROC-AUC are threshold-independent and match the main benchmark, while recall and pLead describe the operating point induced by the fixed threshold. Main-paper Table~\ref{tab:fixed_threshold_transfer} reports the corresponding fixed-threshold evaluation on new routes.

\begin{table*}[!b]
\centering
\caption{Validation-fixed benchmark comparison (additional experiment A). Thresholds selected on source validation routes are held fixed on test; FPR is observed, not constrained to equal 5\%. The AUC columns match the threshold-independent main-paper results.}
\label{tab:validation_fixed_benchmark}
\small
\setlength{\tabcolsep}{3pt}
\begin{tabular*}{\textwidth}{@{\extracolsep{\fill}}llcccccc@{}}
\toprule
Policy & Method & PR-AUC & ROC-AUC & FPR (\%) & $R_F$ & $R_E$ & pLead (s) \\
\midrule
DriveTransformer & Temporal Consistency & 0.127 & 0.715 & 4.30 & 0.126 & 0.601 & 0.705 \\
DriveTransformer & Task-Aware Risk & 0.088 & 0.570 & 4.03 & 0.029 & 0.318 & 0.388 \\
DriveTransformer & TRFD & 0.080 & 0.496 & 2.40 & 0.029 & 0.282 & 0.278 \\
DriveTransformer & SRP & 0.261 & 0.833 & 5.20 & 0.256 & 0.437 & 0.619 \\
\midrule
UniAD & Temporal Consistency & 0.167 & 0.751 & 4.79 & 0.173 & 0.557 & 0.726 \\
UniAD & Task-Aware Risk & 0.070 & 0.506 & 6.76 & 0.065 & 0.479 & 0.659 \\
UniAD & TRFD & 0.090 & 0.520 & 5.26 & 0.107 & 0.495 & 0.684 \\
UniAD & SRP & 0.339 & 0.868 & 5.65 & 0.451 & 0.765 & 1.158 \\
\midrule
VAD & Temporal Consistency & 0.131 & 0.810 & 4.35 & 0.245 & 0.698 & 1.072 \\
VAD & Task-Aware Risk & 0.055 & 0.677 & 3.29 & 0.011 & 0.206 & 0.288 \\
VAD & TRFD & 0.060 & 0.639 & 4.15 & 0.165 & 0.728 & 0.945 \\
VAD & SRP & 0.320 & 0.853 & 5.77 & 0.639 & 0.812 & 1.397 \\
\bottomrule
\end{tabular*}
\end{table*}

\noindent\textbf{Coverage and lead-time trade-offs.}
Observed FPR ranges from 2.40\% to 6.76\%. SRP has the highest frame recall across the three policies and the highest event recall and pLead on UniAD and VAD. On DriveTransformer, temporal consistency has greater event coverage and lead (0.601 and 0.705 seconds) than SRP (0.437 and 0.619 seconds), at observed FPRs of 4.30\% and 5.20\%, respectively. The fixed-threshold results thus qualify, rather than reproduce uniformly, the event-level ordering of the diagnostic benchmark.

\noindent\textbf{Frame labels and metric populations.}
Frame-level metrics use the valid logged policy steps on the evaluated routes. Positive labels are the union of the target event windows, each extending backward from and including its event frame. Overlapping windows assign a single positive label to a frame. Routes without a target event contribute to frame-level evaluation; event-level metrics use the positive routes.

\noindent\textbf{Event and time definitions.}
Each positive route contributes its first event of the evaluated label, giving $N$ selected events. Event recall is the fraction of these events with a qualifying alert. For event $i$, an alert qualifies when $t_i^{\mathrm{evt}}-W_{\mathrm{eval}}\le t<t_i^{\mathrm{evt}}$. Let $t_i^{\mathrm{evt}}$ and $t_i^{\mathrm{alert}}$ denote the event step and the earliest qualifying alert step. At 20 Hz, the penalized lead for event $i$ is
\begin{equation}
\Delta_i=
\begin{cases}
\dfrac{t_i^{\mathrm{evt}}-t_i^{\mathrm{alert}}}{20}, & \text{if alerted},\\[4pt]
0, & \text{otherwise},
\end{cases}
\end{equation}
and the mean penalized lead time is
\begin{equation}
\mathrm{pLead}=\frac{1}{N}\sum_{i=1}^{N}\Delta_i.
\end{equation}
Missed events contribute zero, so pLead reflects both coverage and successful-alert timing. A 40-step window is 2 seconds. Frame-level AUC uses the event-window labels, including event steps; event recall and pLead credit strictly pre-event alerts.

\section{Logistic-regression Baseline Variants}
\label{sec:supp_logreg}

Table~\ref{tab:logreg_comparison} complements the runtime-score benchmark with learned logistic-regression (LR) variants associated with TC, TAR, and TRFD. These variants classify frames from the corresponding monitoring features, whereas the main benchmark evaluates the runtime-score adaptations. Thresholded metrics use the same test-selected diagnostic protocol under empirical frame-level FPR$\le$5\%. SRP-full rows are reproduced for comparison.

\begin{table*}[!t]
\centering
\caption{Logistic-regression variants on Bench2Drive test routes (2-s windows); LR denotes logistic regression. Diagnostic thresholds are selected on test under empirical frame-level FPR$\le$5\%. SRP full combines source-specific readouts with auxiliary event supervision.}
\label{tab:logreg_comparison}
\small
\setlength{\tabcolsep}{3pt}
\begin{tabular*}{\textwidth}{@{\extracolsep{\fill}}llccccc@{}}
\toprule
Policy & Method & PR-AUC & ROC-AUC & $R_F$@5\%FPR & $R_E$@5\%FPR & pLead@5\%FPR (s) \\
\midrule
DriveTransformer & TC (LR) & 0.231 & 0.791 & 0.242 & 0.528 & 0.687 \\
DriveTransformer & TAR (LR) & 0.107 & 0.619 & 0.080 & 0.375 & 0.463 \\
DriveTransformer & TRFD (LR) & 0.057 & 0.520 & 0.020 & 0.200 & 0.312 \\
DriveTransformer & SRP full & 0.261 & 0.833 & 0.300 & 0.497 & 0.701 \\
\midrule
UniAD & TC (LR) & 0.218 & 0.749 & 0.242 & 0.526 & 0.808 \\
UniAD & TAR (LR) & 0.082 & 0.525 & 0.062 & 0.316 & 0.421 \\
UniAD & TRFD (LR) & 0.080 & 0.487 & 0.087 & 0.500 & 0.542 \\
UniAD & SRP full & 0.339 & 0.868 & 0.439 & 0.725 & 1.124 \\
\midrule
VAD & TC (LR) & 0.257 & 0.790 & 0.479 & 0.656 & 1.203 \\
VAD & TAR (LR) & 0.150 & 0.733 & 0.182 & 0.594 & 0.722 \\
VAD & TRFD (LR) & 0.159 & 0.744 & 0.231 & 0.594 & 0.688 \\
VAD & SRP full & 0.320 & 0.853 & 0.559 & 0.713 & 1.229 \\
\bottomrule
\end{tabular*}
\end{table*}

\noindent\textbf{Results.}
TC (LR) improves PR-AUC over the runtime TC adaptation on all three policies. For each policy, SRP full has higher PR-AUC, ROC-AUC, frame recall, and pLead than the reported LR variants. On DriveTransformer, event recall is 0.497 for SRP full and 0.528 for TC (LR). The pLead differences over TC (LR) are 0.014 seconds on DriveTransformer and 0.026 seconds on VAD; these small differences describe the reported point estimates.

\noindent\textbf{Complementary comparisons.}
The runtime-score benchmark and LR variants provide complementary comparisons. The former evaluates runtime-score adaptations, while the latter evaluates learned frame classification from the associated monitoring features. Cross-method rows compare complete monitor designs, with SRP full combining source-specific directions, LME pooling, and auxiliary event supervision.

\section{Implementation and Training Procedure}
\label{sec:supp_implementation}

\noindent\textbf{Representation interface.}
The model-specific exporters map the frozen policies' internal representations to actor, ego, and map source bags. DriveTransformer exposes queries from its actor, ego, and map branches. UniAD supplies actor and ego tracking queries together with lane queries, while VAD supplies agent queries, the ego--map planning representation, and map queries. The source names identify shared semantic roles across these architectures. Separate linear directions operate in each source's feature space, followed by tokenwise scoring and pooling. Training updates the source directions and auxiliary-head parameters while keeping the driving policy and exporter fixed.

\noindent\textbf{Training setup.}
All policies use disjoint 132/44/44 train/validation/test splits of the original Bench2Drive routes. Probes use AdamW \cite{loshchilov2019decoupled}, 50 epochs, learning rate $10^{-3}$, weight decay $10^{-4}$, batch size 128, and weighted binary cross-entropy. The default temperature is $\tau=1$, and full SRP uses auxiliary weight $\lambda=0.5$. Generic validation PR-AUC selects the checkpoint.

\noindent\textbf{Normalization and packed computation.}
Source features are standardized using training-route means and standard deviations:
\begin{equation}
\label{eq:supp_normalization}
\tilde{x}_{t,n}^{s}=\frac{x_{t,n}^{s}-\mu_s}{\max(\sigma_s,\epsilon)},
\qquad \epsilon=10^{-6},
\end{equation}
where division and the maximum are elementwise. These statistics are reused unchanged for validation, testing, and transfer. Tokens are packed with their frame indices, so each pooling segment contains the available tokens from its own frame. LME uses max-subtracted exponentials for numerical stability.

\noindent\textbf{Loss implementation.}
Weighted binary cross-entropy is applied directly to the generic and auxiliary logits. Positive-class weights are estimated separately for the generic target and each event type as the ratio of negative to positive training examples, with the positive count bounded below by one. Auxiliary losses are averaged across the three event types before multiplication by $\lambda$, as in main-paper Eq.~\eqref{eq:srp_loss}. Sigmoid activation converts logits to scores for evaluation and online alerting.

\subsection{Training and Online Inference}
\label{sec:supp_algorithm}

Algorithm~\ref{alg:structured_probe} uses the main paper's notation and equation numbers. During training, within-source pooling supplies the three auxiliary heads, and pooling over all source tokens produces the generic warning score.

\noindent\textbf{Online inference.}
At each policy step, SRP reads the current actor, ego, and map representations from the frozen driving policy and computes the generic risk score online. Training-set normalization, source-specific linear projections, and cardinality-normalized LME pooling produce $r_t$. An alert is issued when $r_t\ge\vartheta$, using the threshold selected on source-validation routes. The generic score alone determines the alert, and auxiliary event-type scores are optional diagnostic outputs. The driving policy remains unchanged.

\begin{algorithm*}[t]
\caption{Structured representation probing with auxiliary supervision by event type}
\label{alg:structured_probe}
\small
\begin{algorithmic}[1]
\REQUIRE Training representations $X_t^s$, aligned failure events, window $W$; frozen policy $f_\theta$; source statistics $(\mu_s,\sigma_s)$; alert threshold $\vartheta$
\ENSURE Probe parameters $\phi$ and online alerts
\item[] \textbf{Training}
\STATE Construct generic and event type window labels $y_t^{(W)}$, $y_{t,c}^{(W)}$
\STATE Initialize source directions and auxiliary event heads
\FOR{each training batch}
  \STATE Normalize source features using $(\mu_s,\sigma_s)$
  \STATE Project elements onto separate source risk directions to obtain $\ell_{t,n}^s$ by Eq.~\eqref{eq:srp_element_logits}
  \STATE Pool within each source to obtain $\rho_t^s$ for auxiliary heads by Eq.~\eqref{eq:srp_source_pooling}
  \STATE Pool all tokens to obtain the generic warning logit $g_t$ by Eq.~\eqref{eq:srp_generic_pooling}
  \STATE Compute event type logits $h_{t,c}$ by Eq.~\eqref{eq:srp_typed_logits}
  \STATE Update $\phi$ with the weighted loss in Eq.~\eqref{eq:srp_loss}
\ENDFOR
\item[] \textbf{Inference}
\FOR{each policy step $t$}
  \STATE Read $X_t^s$ from the frozen policy; normalize and compute $\ell_{t,n}^s$ as in training
  \STATE Pool all source tokens by Eq.~\eqref{eq:srp_generic_pooling} to obtain $r_t=\sigma(g_t)$
  \STATE Issue an alert if $r_t\ge\vartheta$
\ENDFOR
\end{algorithmic}
\end{algorithm*}

\noindent\textbf{Readout cost measurement.}
Main-paper Table~\ref{tab:overhead} reports probe parameter counts and the cost of computing the generic score from source features. FLOPs cover normalization, token projection, LME pooling, and sigmoid activation, with one multiply-add counted as two FLOPs. Readout latency is measured on an NVIDIA A100 GPU, and percentages are relative to the policy runtime. The measured component is the generic risk-score readout; its parameterization and computational scaling are detailed in Sec.~\ref{sec:supp_properties}.

\section{Additional Quantitative Analysis}
\label{sec:supp_quantitative}

\subsection{Window Sensitivity}
\label{sec:supp_window}

\begin{table*}[t]
\caption{Evaluation-window sensitivity on Bench2Drive test routes. Probes are trained with a 2-second window and evaluated at 1, 2, and 3 seconds. Thresholded metrics use test-selected diagnostic thresholds under empirical frame-level FPR$\le$5\%.}
\label{tab:cross_model}
\centering
\small
\setlength{\tabcolsep}{3pt}
\begin{tabular*}{\textwidth}{@{\extracolsep{\fill}}lcccccc@{}}
\toprule
Model & Eval. $W$ & PR-AUC & ROC-AUC & $R_F$@5\%FPR & $R_E$@5\%FPR & pLead@5\%FPR (s) \\
\midrule
DriveTransformer & 20 & 0.176 & 0.856 & 0.320 & 0.437 & 0.376 \\
DriveTransformer & 40 & 0.261 & 0.833 & 0.300 & 0.497 & 0.701 \\
DriveTransformer & 60 & 0.315 & 0.803 & 0.265 & 0.556 & 0.984 \\
\midrule
UniAD & 20 & 0.226 & 0.874 & 0.439 & 0.635 & 0.541 \\
UniAD & 40 & 0.339 & 0.868 & 0.439 & 0.725 & 1.124 \\
UniAD & 60 & 0.408 & 0.853 & 0.429 & 0.804 & 1.711 \\
\midrule
VAD & 20 & 0.226 & 0.855 & 0.571 & 0.643 & 0.564 \\
VAD & 40 & 0.320 & 0.853 & 0.559 & 0.713 & 1.229 \\
VAD & 60 & 0.379 & 0.838 & 0.525 & 0.711 & 1.674 \\
\bottomrule
\end{tabular*}
\end{table*}

\begin{figure*}[t]
\centering
\includegraphics[width=\textwidth]{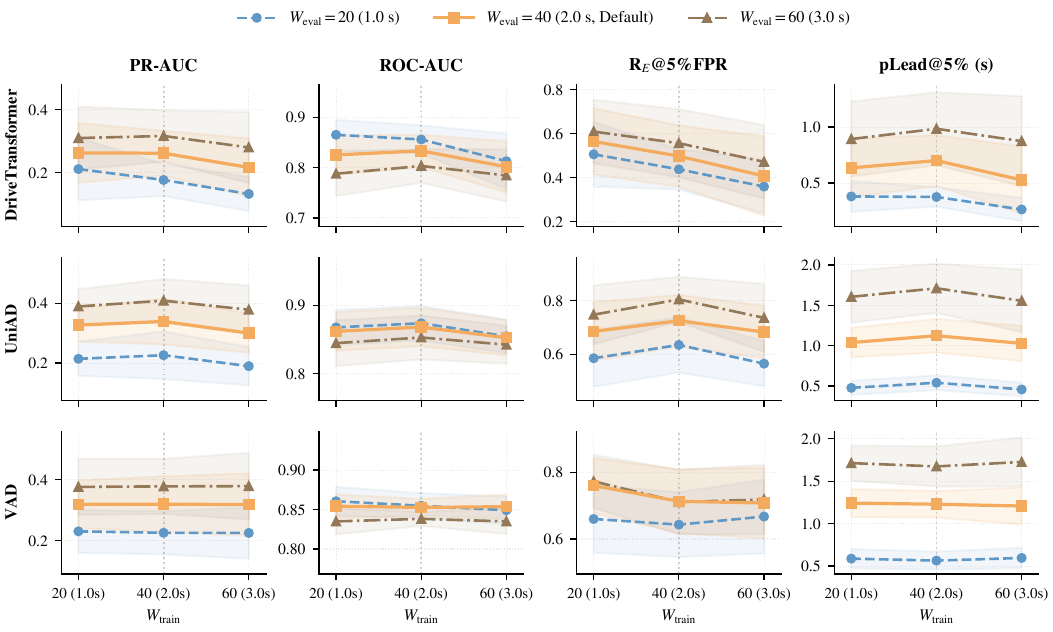}
\caption{Training-window sensitivity across three policies. Rows identify policies, columns show PR-AUC, ROC-AUC, event recall, and pLead, and colors identify evaluation windows. Shaded bands show reported performance variability.}
\label{fig:window_seed_robustness}
\end{figure*}

Table~\ref{tab:cross_model} varies the evaluation window with $W_{\mathrm{train}}=40$, while Figure~\ref{fig:window_seed_robustness} varies the training window at fixed evaluation windows on Bench2Drive \cite{jia2024bench2drive}. Extending the evaluation window raises PR-AUC and pLead but lowers ROC-AUC across the three policies. A longer window changes positive-frame prevalence and the alerts credited as early warnings; PR-AUC depends on prevalence \cite{saito2015precision}, and pLead reflects both the expanded window and coverage. These are different anticipation tasks evaluated with a fixed trained readout, not a uniform model improvement.

Training-window effects differ by policy. DriveTransformer has lower discrimination and event recall at the longest tested training window; UniAD generally peaks near 2 seconds; VAD's PR-AUC and ROC-AUC are comparatively stable. Thus a longer supervision window does not uniformly improve warning.

\subsection{Temperature Sensitivity}
\label{sec:supp_temperature}

\begin{figure}[t]
\centering
\includegraphics[width=\linewidth]{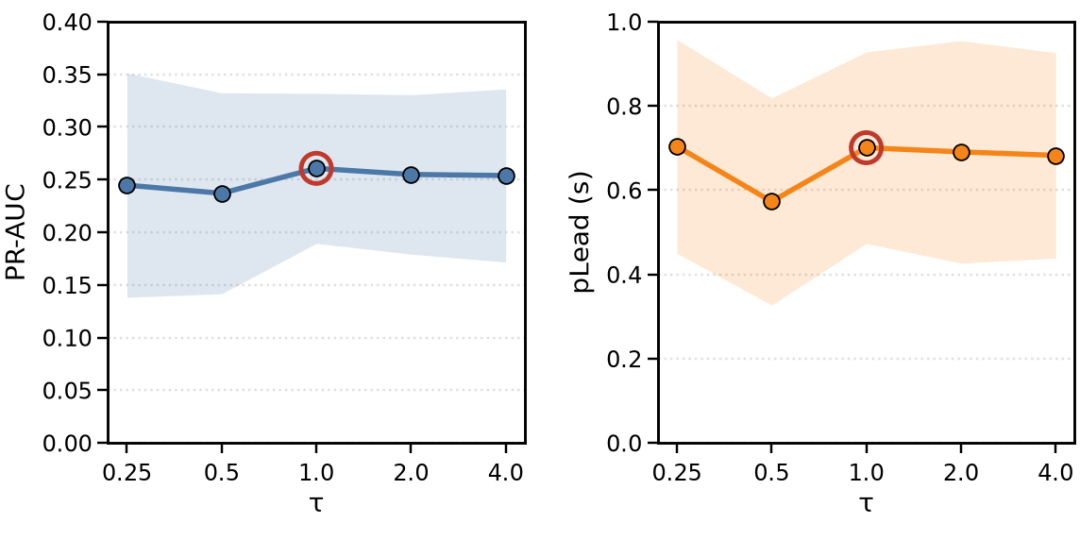}
\caption{Pooling-temperature sensitivity on DriveTransformer. Red circles mark the default $\tau=1$; shaded bands show reported performance variability.}
\label{fig:temperature_sensitivity}
\end{figure}

SRP is retrained at each tested temperature. The default $\tau=1$ has the highest mean PR-AUC, with pLead close to $\tau=0.25$ (0.701 versus 0.704 seconds). PR-AUC and pLead are similar at $\tau=1,2,4$, while $\tau=0.5$ has lower pLead. These results characterize the tested settings rather than establish a unique optimal temperature.

\subsection{Risk Logit Distributions}
\label{sec:supp_logits}

Risk-logit distributions are shown in main-paper Figure~\ref{fig:risk_logit_dist}. Solid and dashed curves distinguish positive and negative event-window frames, and dotted lines locate the policy-specific diagnostic thresholds. These distributions support within-policy discrimination analysis rather than absolute cross-policy calibration.

\subsection{Auxiliary Event-head Diagnostics}
\label{sec:supp_events}

\begin{table*}[t]
\caption{Auxiliary-head diagnostics by event type on Bench2Drive test routes (2-s window). Each head is evaluated against its own labels with a test-selected diagnostic threshold under empirical FPR$\le$5\%; lead times are in seconds. DriveTrans. denotes DriveTransformer.}
\label{tab:typed_attribution}
\centering
\small
\setlength{\tabcolsep}{3pt}
\begin{tabular*}{\textwidth}{@{\extracolsep{\fill}}llccccc@{}}
\toprule
Model & Type & PR-AUC & ROC-AUC & $R_F$@5\%FPR & $R_E$@5\%FPR & pLead@5\%FPR (s) \\
\midrule
DriveTrans. & dynamic & 0.234 & 0.810 & 0.293 & 0.550 & 0.775 \\
DriveTrans. & static & 0.119 & 0.808 & 0.283 & 0.675 & 1.153 \\
DriveTrans. & rule & 0.130 & 0.832 & 0.388 & 0.857 & 1.329 \\
\midrule
UniAD & dynamic & 0.166 & 0.798 & 0.313 & 0.609 & 1.004 \\
UniAD & static & 0.128 & 0.831 & 0.452 & 0.750 & 1.188 \\
UniAD & rule & 0.044 & 0.815 & 0.440 & 0.500 & 0.875 \\
\midrule
VAD & dynamic & 0.208 & 0.827 & 0.502 & 0.612 & 1.193 \\
VAD & static & 0.206 & 0.825 & 0.519 & 0.867 & 1.429 \\
VAD & rule & 0.196 & 0.852 & 0.608 & 0.856 & 1.671 \\
\bottomrule
\end{tabular*}
\end{table*}

The event type with the highest auxiliary-head recall differs by policy: rule violations lead on DriveTransformer, and static collisions lead on UniAD and VAD. On VAD, static collisions have the highest event recall (0.867), whereas rule violations have the longest pLead (1.671 seconds). These head-specific results are distinct from the generic alarm applied to route groups in Sec.~\ref{sec:supp_qualitative}.

\subsection{Source Deletion at Inference}
\label{sec:supp_deletion}

\begin{table}[t]
\centering
\caption{Fixed-probe source deletion on new Bench2Drive routes. Crosses mark removed sources. Each setting reselects a target-set diagnostic threshold under empirical FPR$\le$5\%; lead times are in seconds.}
\label{tab:source_deletion_v2}
\small
\setlength{\tabcolsep}{2pt}
\begin{tabular*}{\columnwidth}{@{\extracolsep{\fill}}lccccc@{}}
\toprule
Setting & \shortstack{PR-\\AUC} & \shortstack{ROC-\\AUC} & \shortstack{$R_F$\\@5\%} & \shortstack{$R_E$\\@5\%} & \shortstack{pLead\\@5\%} \\
\midrule
Full & 0.363 & 0.807 & 0.360 & 0.681 & 1.029 \\
\textcolor{red}{$\times$} actor & 0.336 & 0.785 & 0.326 & 0.631 & 0.964 \\
\textcolor{red}{$\times$} ego & 0.217 & 0.716 & 0.205 & 0.411 & 0.606 \\
\textcolor{red}{$\times$} map & 0.358 & 0.794 & 0.356 & 0.670 & 1.008 \\
\bottomrule
\end{tabular*}
\end{table}

The complete DriveTransformer probe and deletion variants are evaluated on the new 220-route Bench2Drive set.\footnote{Released with Bench2Drive 0.0.4: \url{https://github.com/Thinklab-SJTU/Bench2Drive/blob/0.0.4/leaderboard/data/bench2drive_0.0.4_val.xml}.} Parameters are fixed and generic pooling is renormalized over the remaining tokens. Ego deletion has the largest effect on ranking and event coverage. This is a fitted-input deletion diagnostic, not causal attribution or fixed-threshold robustness. The retrained source ablations use a different dataset and measure attainable performance after fitting a new readout, so their values cannot be directly subtracted from these results.

\section{Additional Qualitative and Temporal Analysis}
\label{sec:supp_qualitative}

\subsection{Generic Warning by Event Group}

\begin{figure}[t]
\centering
\includegraphics[width=\linewidth]{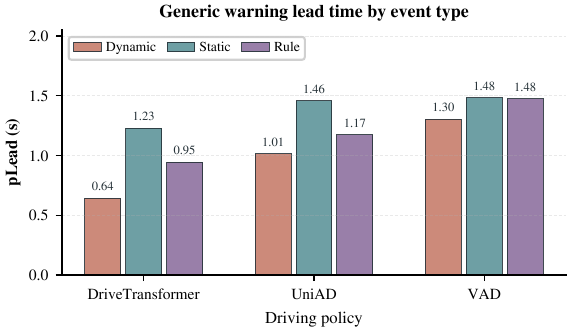}
\caption{Generic-alarm pLead on groups of Bench2Drive test routes containing dynamic collision, static collision, or rule violation. The test-selected diagnostic protocol uses the 5\% frame-level FPR constraint. Groups are described by observed failure types, not auxiliary-head predictions.}
\label{fig:plead_by_type}
\end{figure}

The reported generic alarm has longer mean pLead in route groups containing static collisions or rule violations than in the dynamic-collision group. Because missed events contribute zero, this difference concerns both coverage and warning timing. These route-group results should not be interpreted as auxiliary-head classification or mutually exclusive first-event classes.

\subsection{Temporal Score Profiles}

\begin{figure*}[t]
\centering
\includegraphics[width=\textwidth]{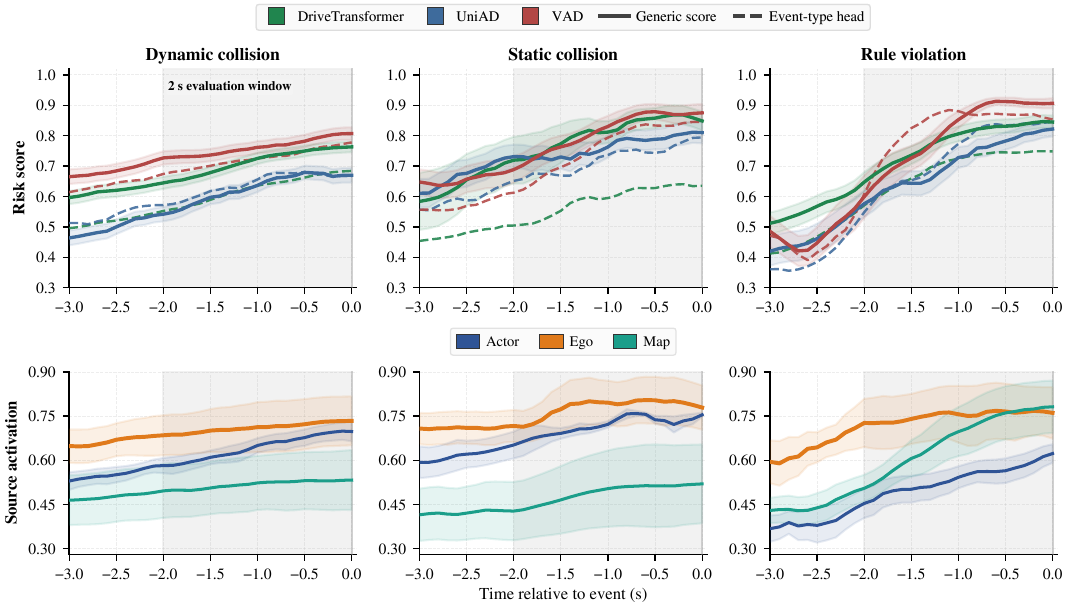}
\caption{Event-aligned temporal profiles on Bench2Drive test routes grouped by observed failure type. Upper curves compare generic and auxiliary scores for each policy; the bottom row averages source scores across policies. Shading marks the 2-second evaluation window. Cross-policy source averages are descriptive, not calibrated importance scores.}
\label{fig:temporal_dyn}
\end{figure*}

Generic scores generally rise toward selected target events. For dynamic collisions, the generic and auxiliary readouts track closely; for static collisions, DriveTransformer's auxiliary score remains below its generic score even as both rise. These readouts share source directions but differ in aggregation and supervision targets.

The source profiles visualize evidence evolution, but separately trained policies need not share logit scales. Consequently, the bottom-row averages do not establish absolute source importance or cross-policy comparability. We avoid causal interpretations and use the controlled source ablations and fixed-probe deletion results for quantitative source analysis.

Policy-specific qualitative cases are shown in main-paper Figure~\ref{fig:policy_warning_cases}. These examples illustrate individual warning traces; overall reliability is assessed using the frame and event metrics.

\subsection{Additional Route Visualizations}
\label{sec:supp_route_visualizations}

\begin{figure*}[p]
\centering
\includegraphics[width=\textwidth]{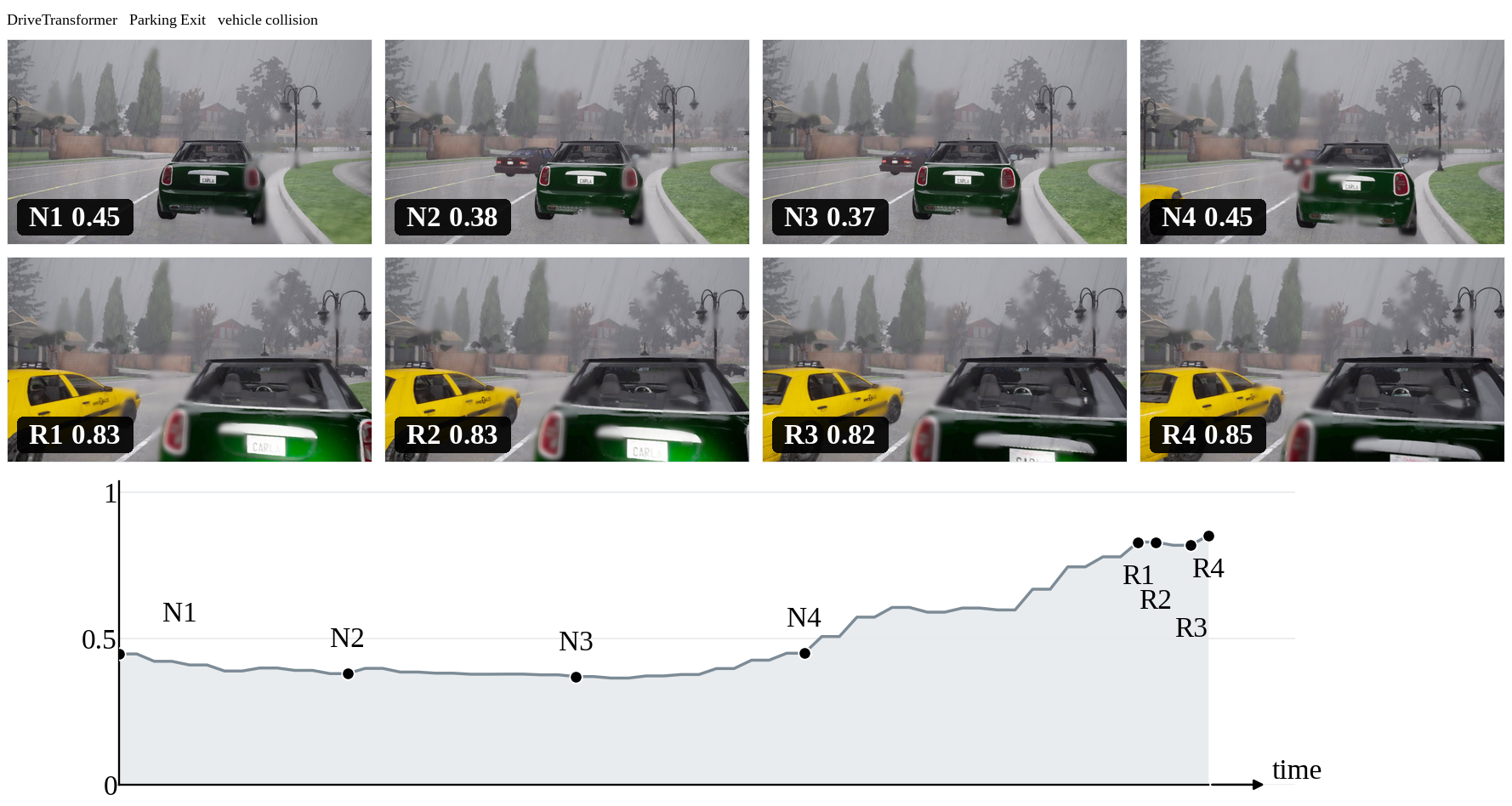}
\caption{DriveTransformer parking-exit sequence with a vehicle collision. Earlier and later snapshots are linked to the generic score trace by the N1--N4 and R1--R4 markers.}
\label{fig:supp_grayblue_dt1}
\end{figure*}

\begin{figure*}[p]
\centering
\includegraphics[width=\textwidth]{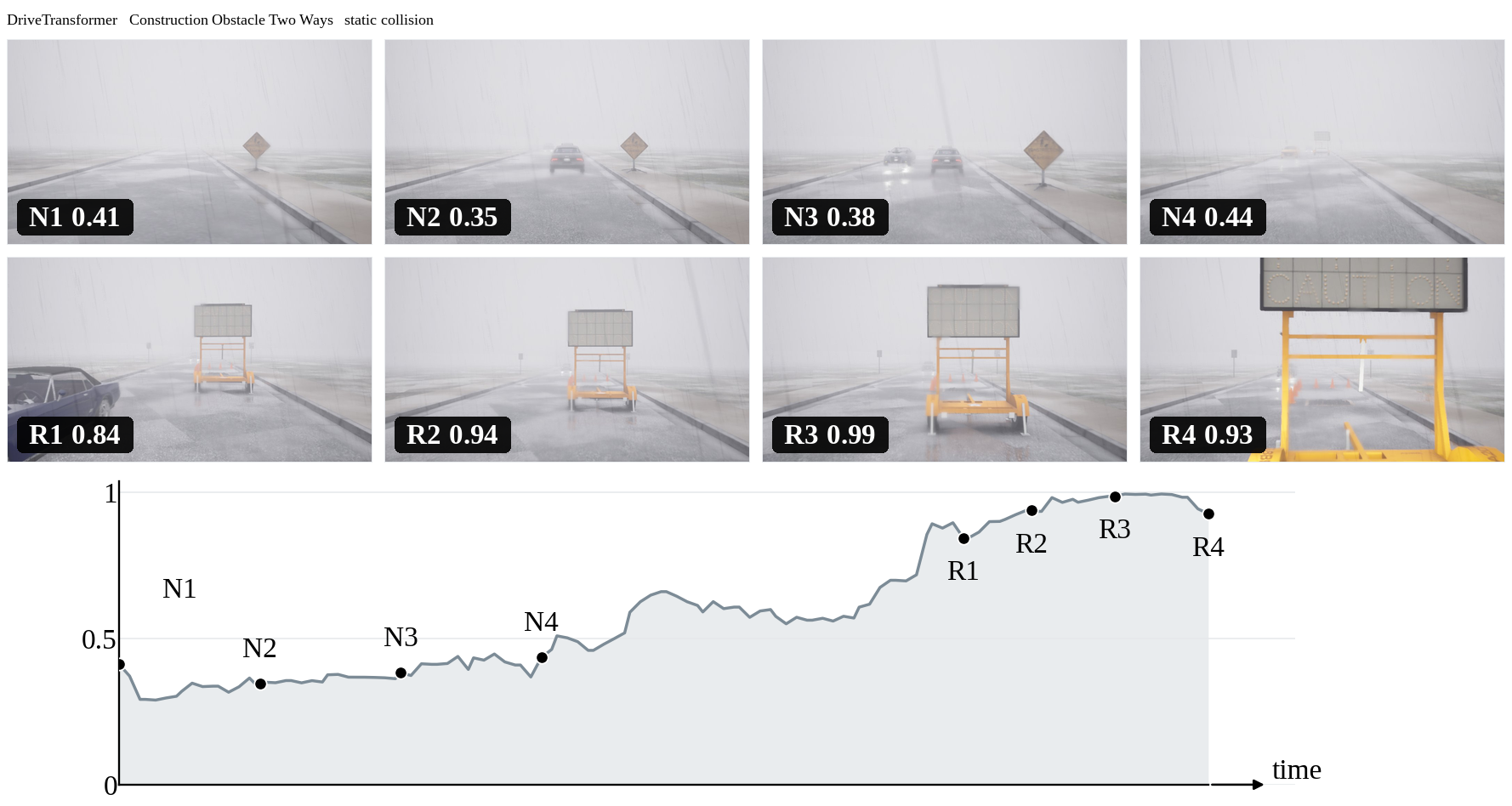}
\caption{DriveTransformer construction-obstacle sequence with a static collision. The later frames show the construction barrier approaching in the front-camera view while the generic score rises.}
\label{fig:supp_grayblue_dt2}
\end{figure*}

\begin{figure*}[p]
\centering
\includegraphics[width=\textwidth]{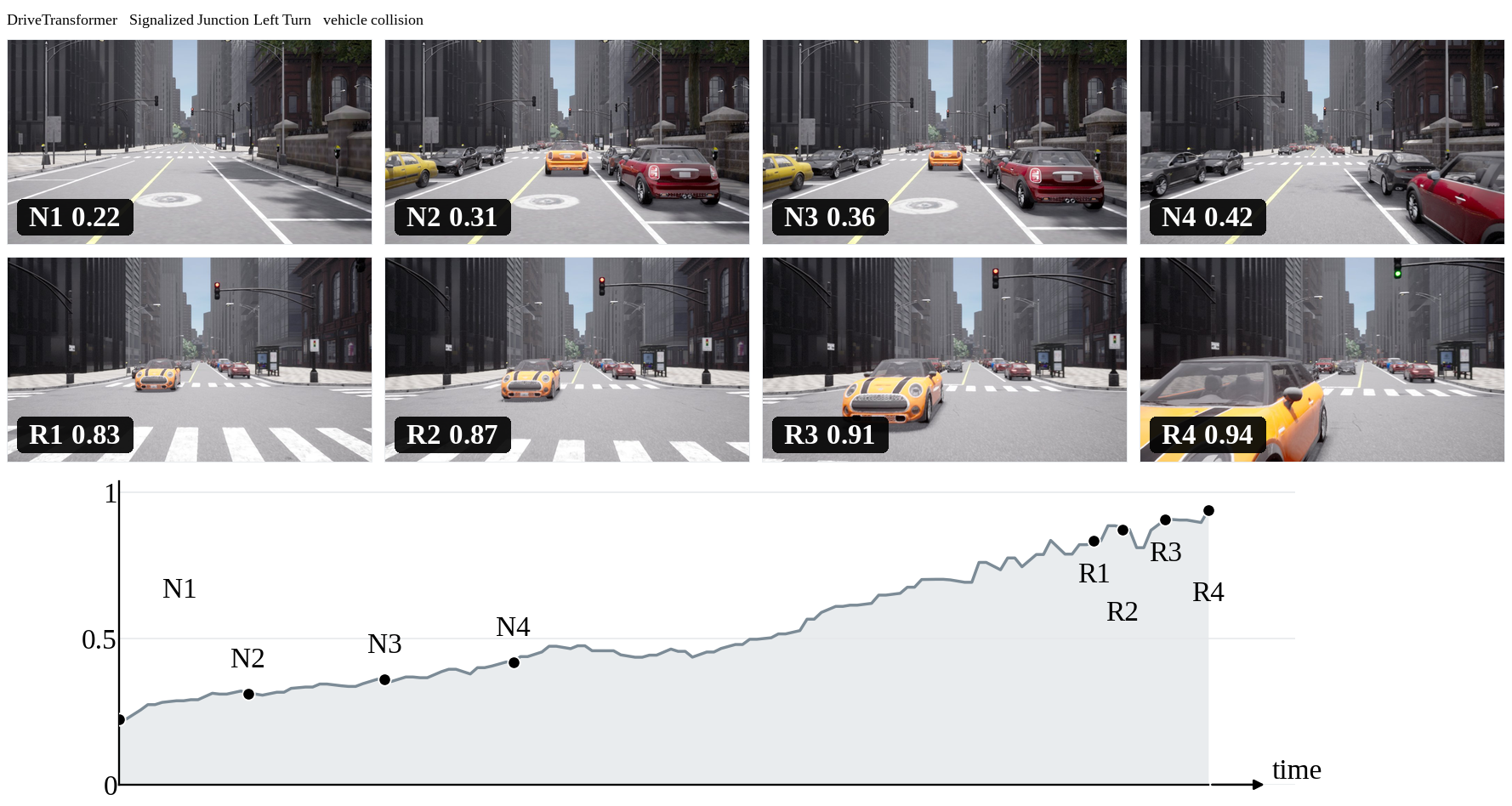}
\caption{DriveTransformer signalized-junction left-turn sequence with a vehicle collision. The yellow vehicle becomes prominent in the later snapshots as the score enters a higher range.}
\label{fig:supp_grayblue_dt3}
\end{figure*}

\begin{figure*}[p]
\centering
\includegraphics[width=\textwidth]{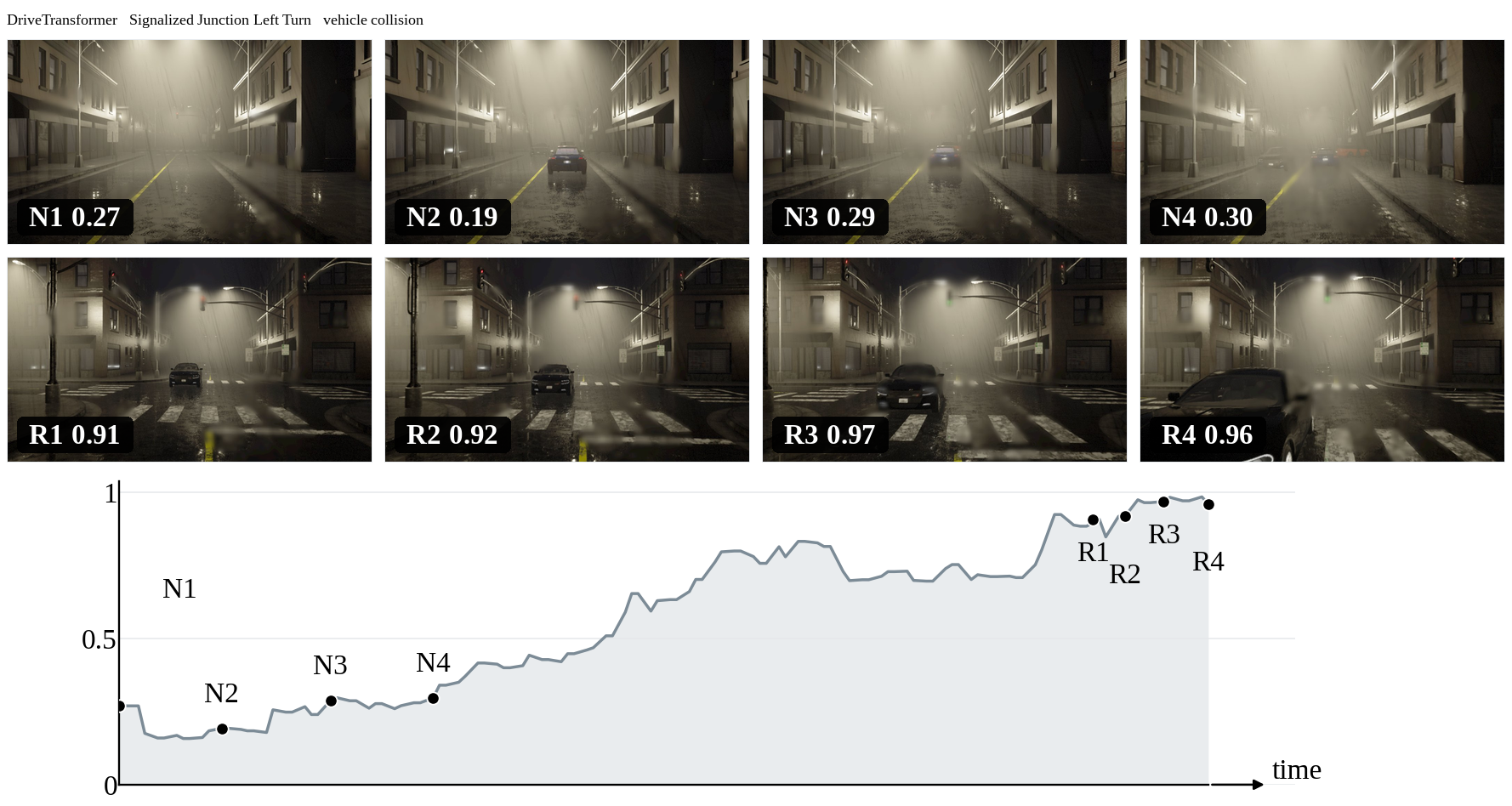}
\caption{DriveTransformer signalized-junction left-turn sequence in low-light conditions, with a vehicle collision. The later snapshots show a nearby vehicle alongside a high-score interval.}
\label{fig:supp_grayblue_dt4}
\end{figure*}

\begin{figure*}[p]
\centering
\includegraphics[width=\textwidth]{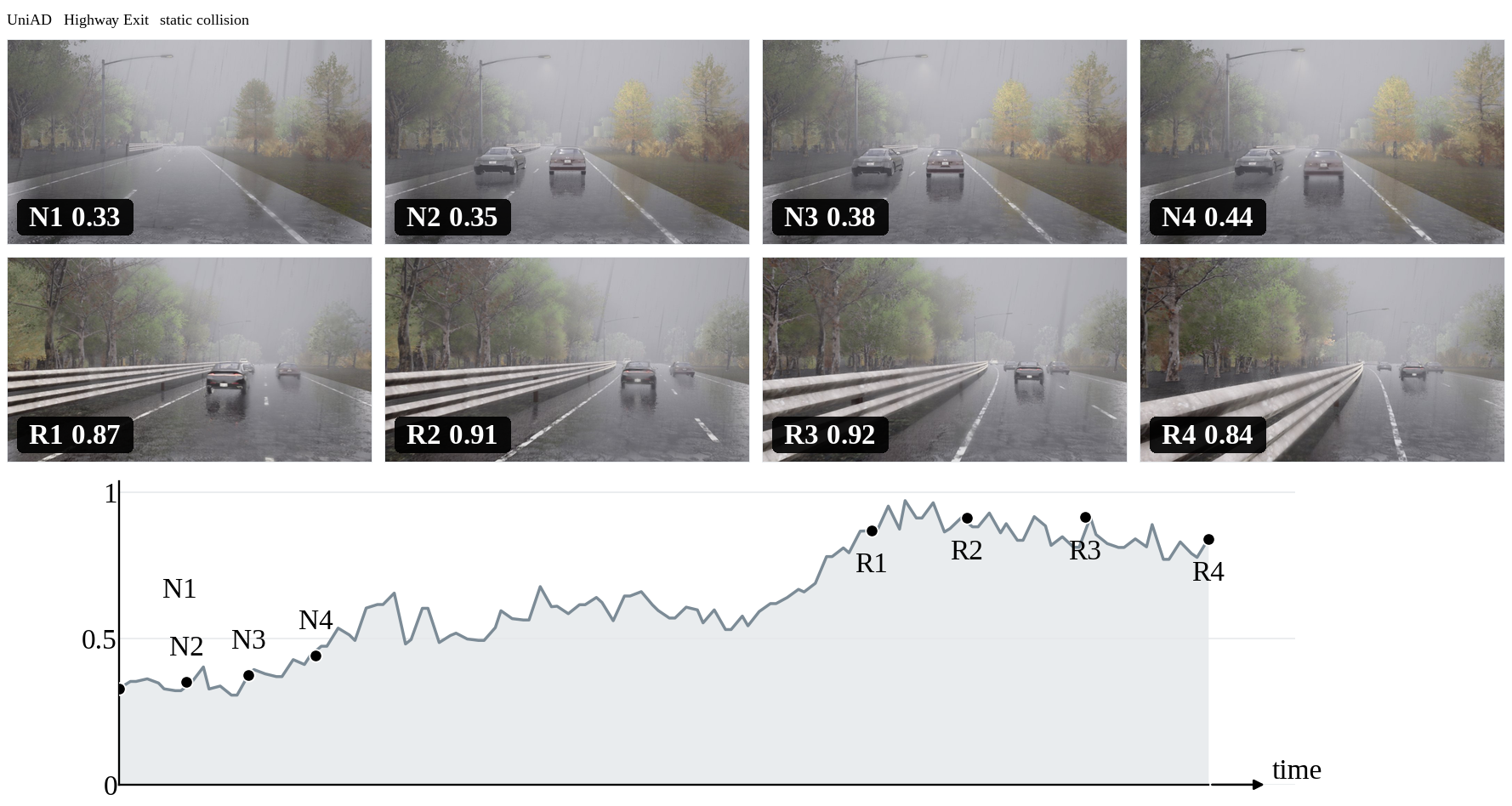}
\caption{UniAD highway-exit sequence with a static collision. The later camera views show the left guardrail at close range, with the marked frames located on the generic score trace.}
\label{fig:supp_grayblue_uniad1}
\end{figure*}

\begin{figure*}[p]
\centering
\includegraphics[width=\textwidth]{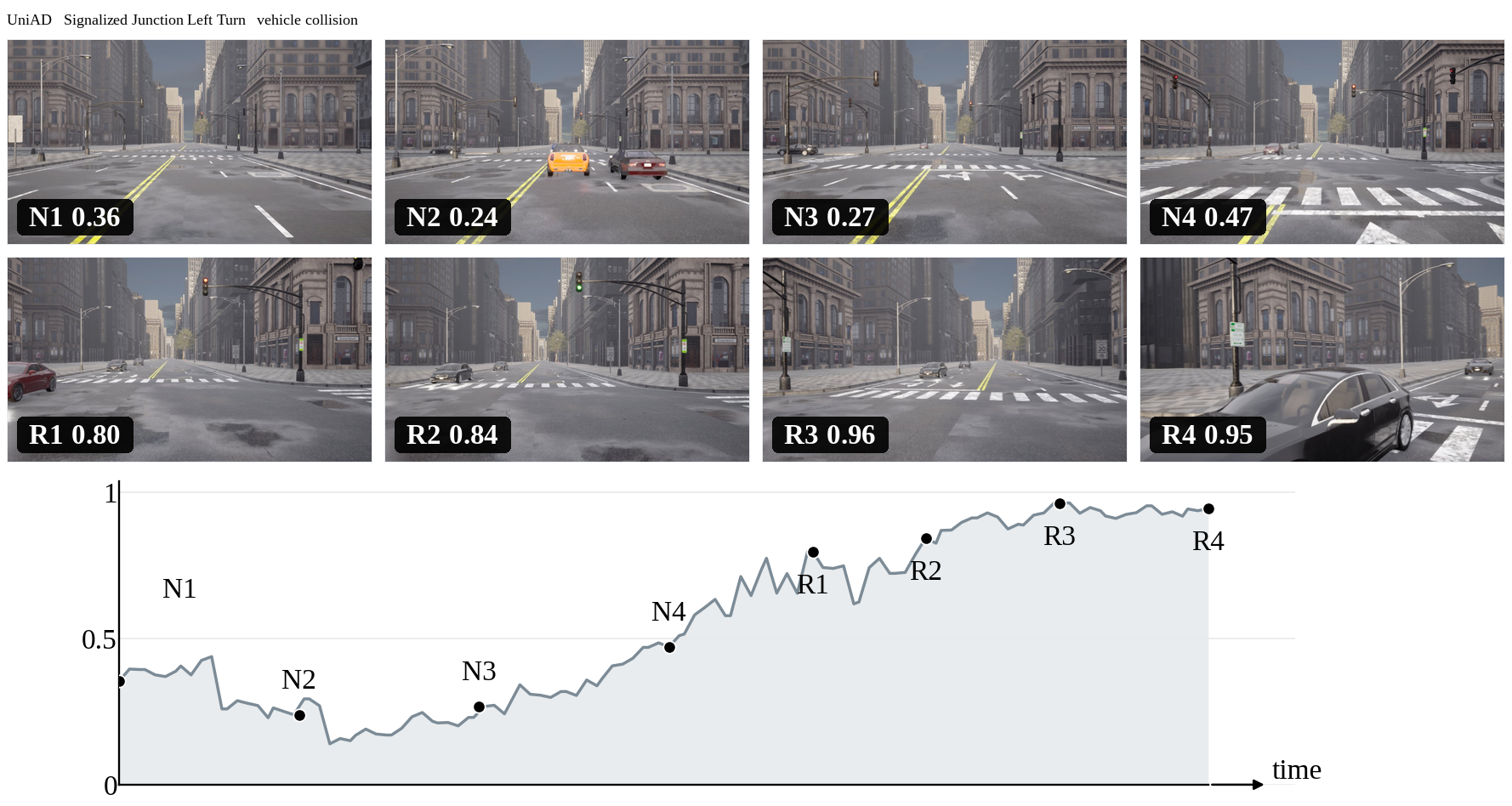}
\caption{UniAD signalized-junction left-turn sequence with a vehicle collision. The black vehicle becomes prominent in the later front-camera snapshots.}
\label{fig:supp_grayblue_uniad2}
\end{figure*}

\begin{figure*}[p]
\centering
\includegraphics[width=\textwidth]{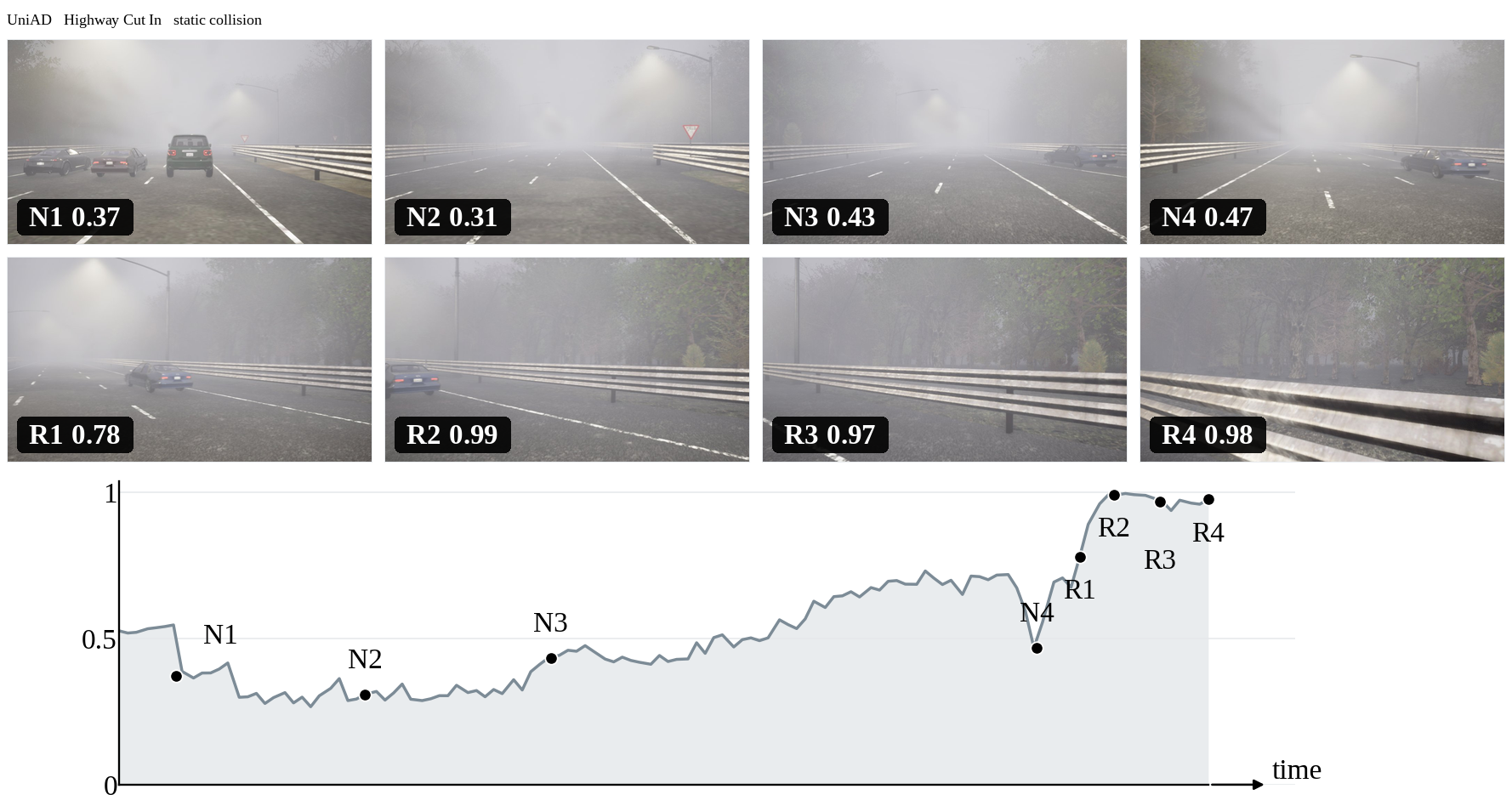}
\caption{UniAD highway cut-in sequence with a static collision. The snapshots link the earlier traffic scene and the later guardrail approach to the generic score trace.}
\label{fig:supp_grayblue_uniad3}
\end{figure*}

\begin{figure*}[p]
\centering
\includegraphics[width=\textwidth]{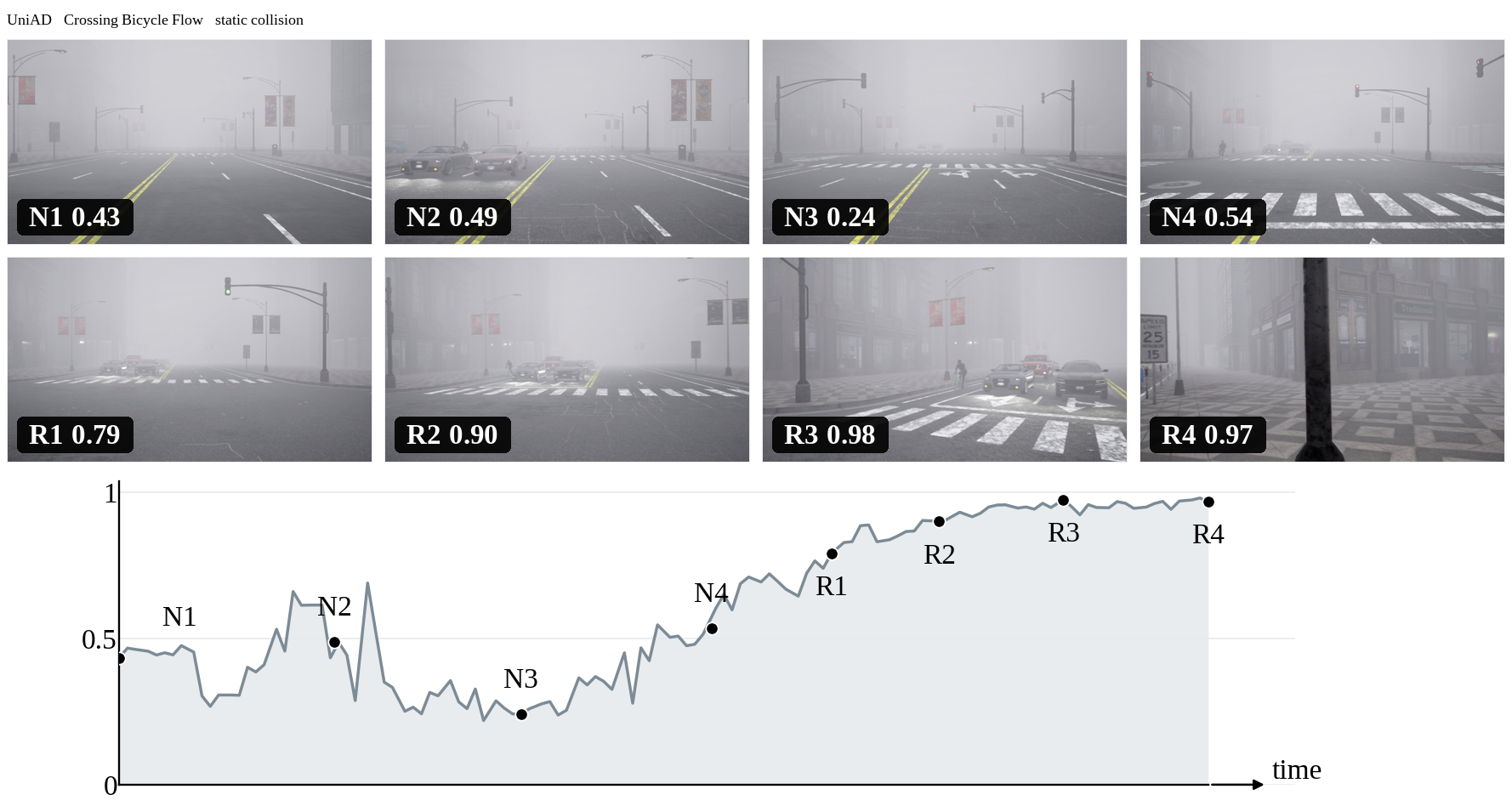}
\caption{UniAD crossing-bicycle-flow sequence with a static collision. The later snapshots follow the junction traversal toward the roadside as the score enters a higher range.}
\label{fig:supp_grayblue_uniad4}
\end{figure*}

\begin{figure*}[p]
\centering
\includegraphics[width=\textwidth]{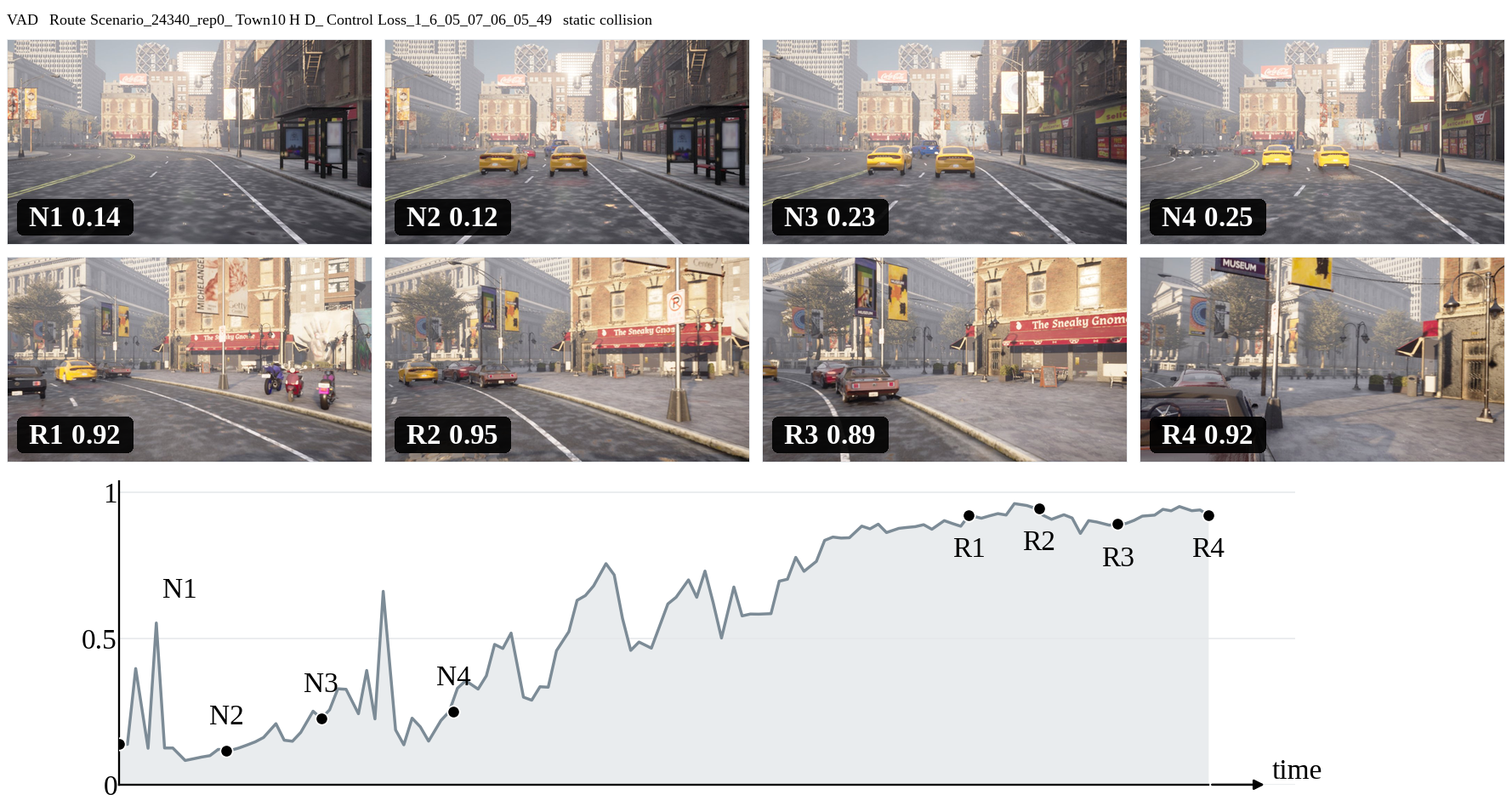}
\caption{VAD urban control-loss sequence with a static collision. The later views approach the roadside; the trace contains fluctuations followed by a high-score interval.}
\label{fig:supp_grayblue_vad1}
\end{figure*}

\begin{figure*}[p]
\centering
\includegraphics[width=\textwidth]{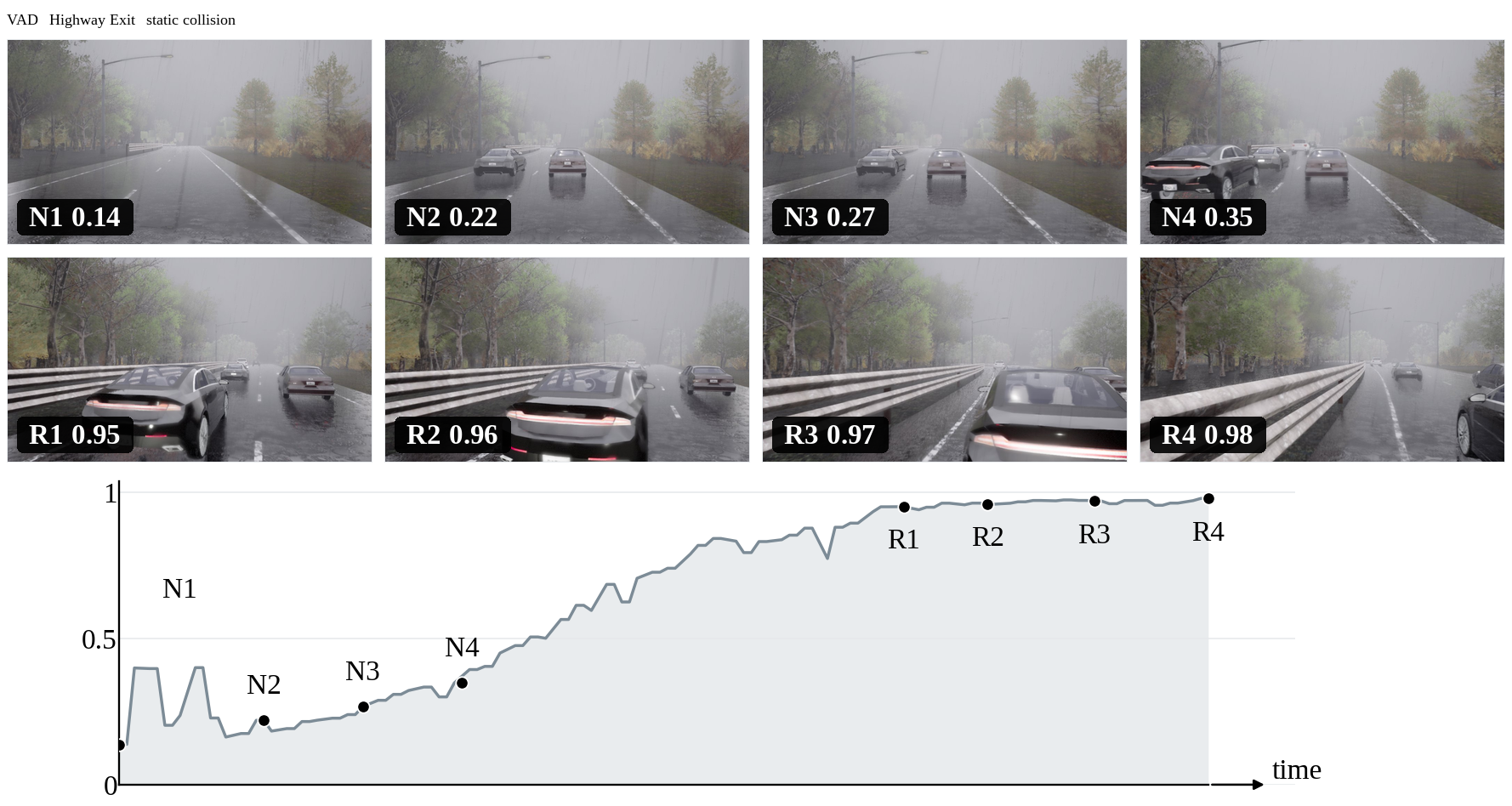}
\caption{VAD highway-exit sequence with a static collision. Nearby traffic and the left guardrail become prominent in the later snapshots.}
\label{fig:supp_grayblue_vad2}
\end{figure*}

\begin{figure*}[p]
\centering
\includegraphics[width=\textwidth]{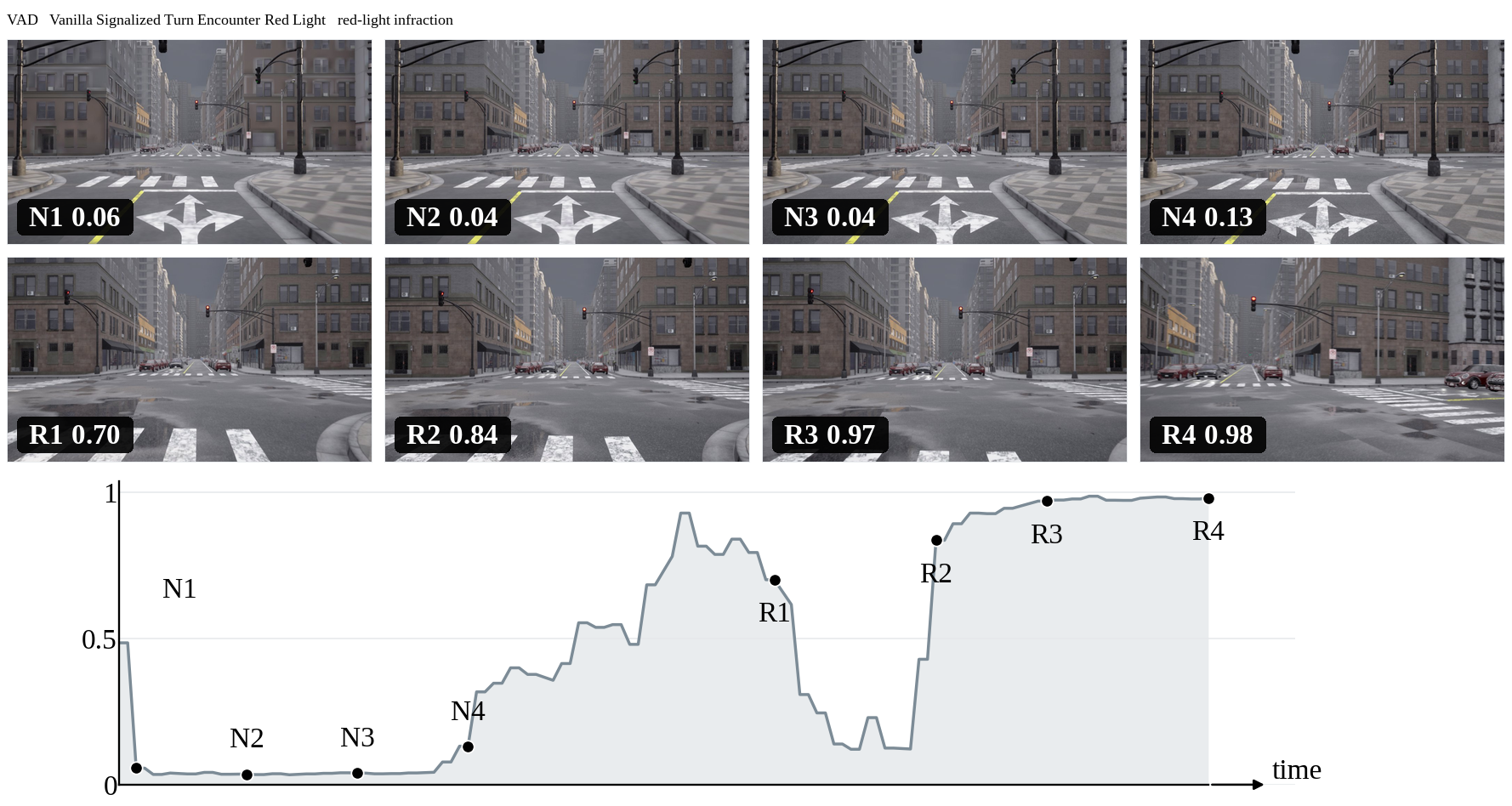}
\caption{VAD signalized-turn sequence with a red-light infraction. The generic score varies during the approach and rises again in the later marked interval.}
\label{fig:supp_grayblue_vad3}
\end{figure*}

\begin{figure*}[p]
\centering
\includegraphics[width=\textwidth]{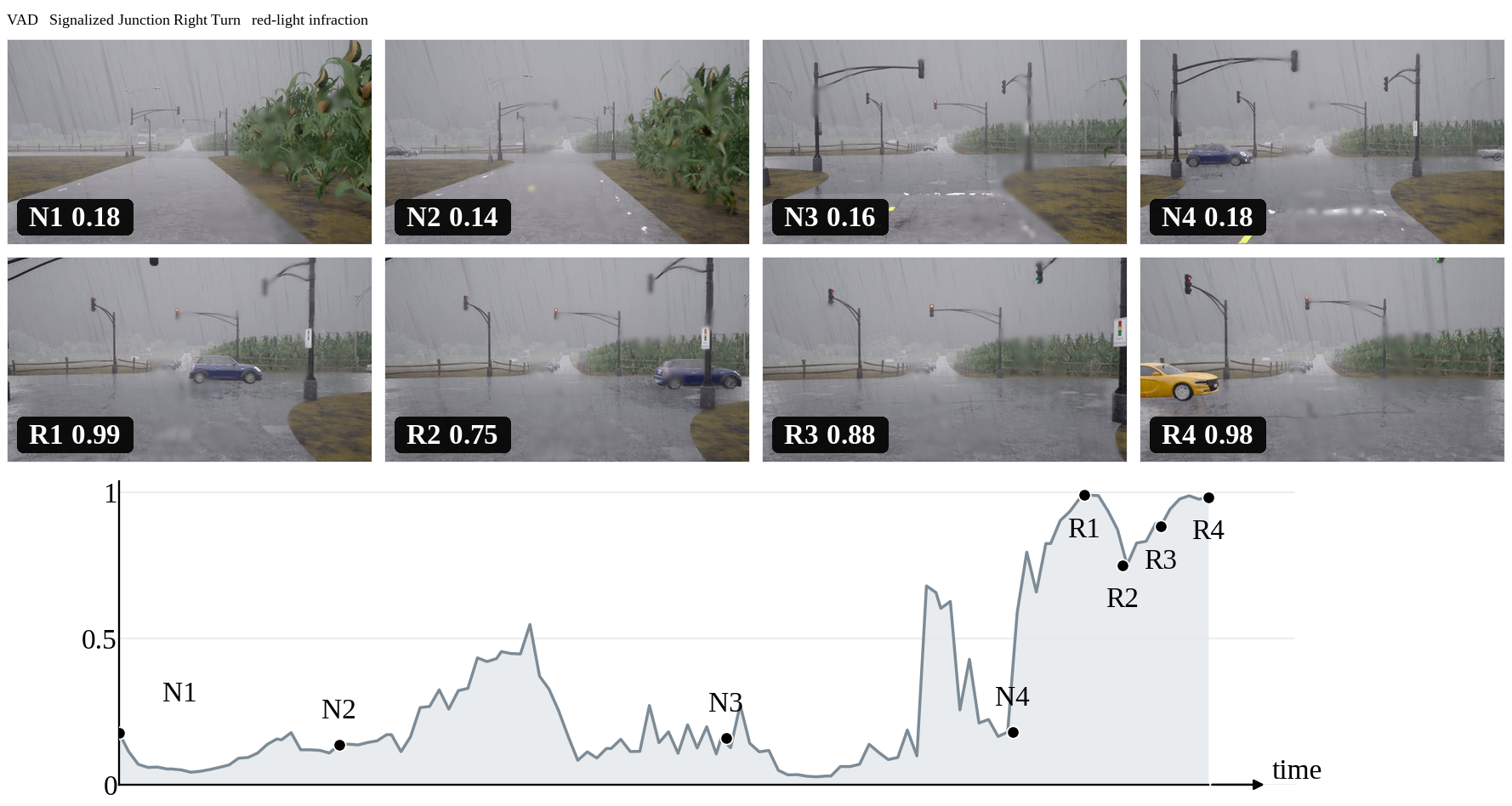}
\caption{VAD right-turn sequence at a signalized junction, with a red-light infraction. The later markers align with a high-score interval near the junction.}
\label{fig:supp_grayblue_vad4}
\end{figure*}

Figures~\ref{fig:supp_grayblue_dt1}--\ref{fig:supp_grayblue_vad4} show twelve additional route visualizations, with four examples for each policy. Each figure combines two rows of front-camera snapshots with the corresponding generic score trace. N1--N4 identify earlier, lower-score snapshots, while R1--R4 identify later, higher-score snapshots. The numbers on the snapshots report the generic scores at the selected frames, and black dots link these frames to their positions on the trace. The gray-blue line follows the score over route time; the light fill shows the area beneath that line.

The DriveTransformer examples in Figures~\ref{fig:supp_grayblue_dt1}--\ref{fig:supp_grayblue_dt4} cover parking exit, construction obstacles, and signalized-junction left turns. Their later snapshots show nearby vehicles or a construction barrier occupying more of the camera view as the traces move toward higher scores. The UniAD examples in Figures~\ref{fig:supp_grayblue_uniad1}--\ref{fig:supp_grayblue_uniad4} include highway exit, highway cut-in, a junction left turn, and crossing bicycle flow. The marked frames connect these traffic scenes and later vehicle or roadside encounters to the changing score traces.

The VAD examples in Figures~\ref{fig:supp_grayblue_vad1}--\ref{fig:supp_grayblue_vad4} include static-collision and red-light-infraction sequences. Their traces contain local fluctuations as well as later high-score intervals. In the signalized-turn example, the score rises during the approach, decreases, and then rises again near the later marked frames. The snapshots make these different stages visible alongside the trace.

\section{Readout Properties and Computational Complexity}
\label{sec:supp_properties}

This section develops the properties of the source-wise and generic pooling operators in main-paper Eqs.~\eqref{eq:srp_source_pooling} and~\eqref{eq:srp_generic_pooling}, together with the computational scaling of the online readout.

\subsection{Pooling Properties}

\noindent\textbf{Bounds and invariance.}
For a nonempty bag of $N$ element logits and temperature $\tau>0$, write its LME score as $\rho$ and let $\ell_{\max}=\max_n\ell_n$. The exponential sum is at least $e^{\ell_{\max}/\tau}$ and at most $N e^{\ell_{\max}/\tau}$, giving
\begin{equation}
\label{eq:supp_pool_bounds}
\ell_{\max}-\tau\log N\le\rho\le\ell_{\max}.
\end{equation}
Jensen's inequality also gives $\bar{\ell}\le\rho$, where $\bar{\ell}=N^{-1}\sum_n\ell_n$. Thus LME lies between the mean and maximum, and the bounds imply $\lim_{\tau\to0^+}\rho=\ell_{\max}$. Permuting the elements preserves both the sum and cardinality. Repeating the entire bag $k$ times multiplies both by $k$, leaving the score unchanged. These properties preserve sensitivity to the logits and bag composition while removing the scaling introduced by uniform replication.

\noindent\textbf{Gradient weighting.}
Differentiating the pooled score gives
\begin{equation}
\label{eq:supp_pool_gradient}
\frac{\partial\rho}{\partial\ell_n}
=\frac{e^{\ell_n/\tau}}{\sum_j e^{\ell_j/\tau}}.
\end{equation}
The derivatives are nonnegative and sum to one. Higher-logit elements receive larger pooling gradient weights, with temperature controlling their concentration. This smooth weighting connects the pooled warning score to salient element evidence during training.

\noindent\textbf{Aggregation across sources.}
Let $\mathcal{S}_t^+=\{s\in\mathcal{S}:N_t^s>0\}$ identify the nonempty sources and define $p_t^s=N_t^s/N_t$, with $N_t>0$. Substituting the source-wise scores into the all-token generic readout yields
\begin{equation}
\label{eq:supp_source_aggregation}
g_t=\tau\log\!\left(
\sum_{s\in\mathcal{S}_t^+}p_t^s e^{\rho_t^s/\tau}
\right).
\end{equation}
This is the same generic score as main-paper Eq.~\eqref{eq:srp_generic_pooling}, expressed through source summaries. It makes the token-proportion weighting explicit. Uniform replication of all source bags preserves both the proportions and source scores. Auxiliary heads instead use the source-score vector in main-paper Eq.~\eqref{eq:srp_typed_logits}.

\subsection{Computational Complexity}

\noindent\textbf{Parameterization.}
Let $d_s$ be the feature dimension of source $s$. Each source contributes one direction and one bias; each auxiliary event head contributes $|\mathcal{S}|$ weights and one bias. The trainable parameter counts are
\begin{equation}
\label{eq:supp_parameter_counts}
\begin{aligned}
P_{\mathrm{gen}}&=\sum_{s\in\mathcal{S}}(d_s+1),\\
P_{\mathrm{aux}}&=|\mathcal{C}|(|\mathcal{S}|+1).
\end{aligned}
\end{equation}
The full probe has $P_{\mathrm{gen}}+P_{\mathrm{aux}}$ trainable parameters. Generic online alerting uses the source directions and biases, while the auxiliary heads support event-type training and optional diagnostics.

\noindent\textbf{Stable pooling and per-frame cost.}
LME is evaluated using the maximum logit $m=\ell_{\max}$:
\begin{equation}
\label{eq:supp_stable_pooling}
\rho=m+\tau\log\!\left(
\frac{1}{N}\sum_{n=1}^{N}e^{(\ell_n-m)/\tau}
\right).
\end{equation}
The shifted exponents are nonpositive, and the expression evaluates the same normalized exponential mean. Normalization and projection scale with the total number of feature coordinates; pooling reduces the $N_t$ element logits. The resulting generic readout cost is
\begin{equation}
\label{eq:supp_readout_complexity}
\mathcal{O}\!\left(\sum_{s\in\mathcal{S}}N_t^s d_s+N_t\right).
\end{equation}
Optional auxiliary scoring reuses these element logits, adding $\mathcal{O}(N_t)$ within-source pooling and $\mathcal{O}(|\mathcal{C}||\mathcal{S}|)$ linear-head computation. This linear per-frame structure characterizes the SRP readout measured in main-paper Table~\ref{tab:overhead}, separately from the frozen driving policy.

\clearpage

\end{document}